\documentclass{article}
\usepackage{colm2024_conference}
\usepackage[T1]{fontenc}
\usepackage{lmodern}
\usepackage{hyperref}
\hypersetup{hypertexnames=false,hyperfootnotes=false}
\usepackage{url}
\usepackage{booktabs}
\usepackage{amsfonts}
\usepackage{amsmath}
\usepackage{amssymb}
\usepackage{microtype}
\usepackage{multirow}
\usepackage{xcolor}
\usepackage{array}
\usepackage{graphicx}
\usepackage{float}
\usepackage{flafter}
\usepackage{wrapfig}
\usepackage{placeins}
\usepackage{etoc}
\usepackage{tcolorbox}
\tcbuselibrary{skins,breakable,listings}
\usepackage[capitalise,noabbrev]{cleveref}

\definecolor{vhdpurple}{RGB}{79,70,229}
\definecolor{vhdpromptbg}{RGB}{249,247,253}
\newtcolorbox{vhdprompt}[1]{
  enhanced,
  breakable,
  title={#1},
  colback=vhdpromptbg,
  colframe=vhdpurple,
  colbacktitle=vhdpurple,
  coltitle=white,
  fonttitle=\bfseries,
  boxrule=0.45pt,
  arc=1mm,
  left=1.5mm,
  right=1.5mm,
  top=1mm,
  bottom=1mm,
  before skip=6pt,
  after skip=6pt
}

\newcommand{\vhd}{VHD-Play}

\makeatletter
\providecommand*{\input@path}{}
\g@addto@macro\input@path{{}{tables/}}
\makeatother

\title{Verifiable Hidden Dynamics Play:\\
Generating Agentic RL Environments from Solved Mechanisms}
\author{
{Xinjie Shen}\textsuperscript{1,2*}, 
Wei Fan\textsuperscript{2}, 
Xudong Guo\textsuperscript{2,\dag}, 
Jianhong Tu\textsuperscript{2},
Yang Su\textsuperscript{2}, 
Chuqiao Kuang\textsuperscript{2}, \\ 
Yinger Zhang\textsuperscript{2},
Lianghao Deng\textsuperscript{2}, 
Dayiheng Liu\textsuperscript{2}\\[0.6em]
\textsuperscript{1}Georgia Institute of Technology \qquad
\textsuperscript{2}Alibaba Token Foundry, Alibaba Group
}

\begin{document}

\etocdepthtag.toc{vhdmain}
\etocsettagdepth{vhdmain}{subsection}
\etocsettagdepth{vhdappendix}{none}

\renewcommand{\thefootnote}{\fnsymbol{footnote}}
\maketitle
\footnotetext[1]{Work done during summer internship. Correspondence: \texttt{xinjie@gatech.edu} \quad \textsuperscript{\dag}Project lead.}

\renewcommand{\thefootnote}{\arabic{footnote}}
\setcounter{footnote}{0}

\begin{abstract}
Language-model agents increasingly face long-horizon tasks with evolving state, interdependent decisions, and delayed outcomes. Scaling their training requires diverse agentic environments, dependable outcome signals, and low extension cost. Existing generation pipelines commonly construct an environment before defining its outcome rule or annotating its trajectories, leaving dynamics and evaluation to be aligned post hoc. \vhd\ reverses this dependency by sampling and solving a mathematical model before a corpus-grounded setter renders its decision process as stateful tools. The executable dynamics and trajectory-scoring reference are inherited from the same solved model. The pipeline produces 3{,}300 diverse agentic environments at a cost of a few cents each. Training Qwen3.6-35B-A3B on three families raises its mean agentic score from 0.204 to 0.815 in a five-family diagnostic. Gains also appear on held-out instances from all three training families and eight unseen mechanism families, then extend beyond the generated substrate to external benchmarks for general function calling, travel planning, and 365-day e-commerce. On E-Commerce Bench, the trained checkpoint completes every run without bankruptcy and exceeds Qwen3.7-Max. We compare written-out problems with stateful versions that reveal or hide their parameters. The comparison shows that most of the learnable gap lies in stateful interaction rather than underlying problem solving. A frozen 35B setter realizes larger environments, and scale-matched training retains gains as mechanism size and horizon grow, indicating the potential for an evolving training substrate.
\end{abstract}

\section{Introduction}
\label{sec:intro}

\begin{figure}[t]
  \centering
  \vspace{-1.0em}
  \includegraphics[width=0.92\linewidth]{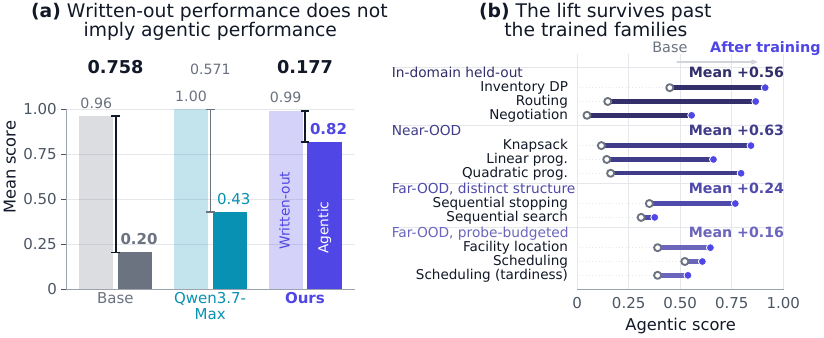}
  \vspace{-0.9em}
  \caption{(a) The same optimization problem appears as written-out QA or embedded in a stateful agentic
  environment, but performance does not transfer automatically. (b) Agentic training improves held-out instances
  from trained families, unseen optimization families (near-OOD), and families with different decision or
  observation structures (far-OOD). DP, LP, and QP denote dynamic, linear, and quadratic programming.}
  \label{fig:teaser}
  \vspace{-1.3em}
\end{figure}

As language-model agents reach broader deployment, they are increasingly expected to complete diverse,
end-to-end workflows rather than answer isolated requests. Emerging applications ask them to repair software
through repeated iterations~\citep{cite_slopcodebench}, resolve customer requests across
tools~\citep{cite_tau_bench}, and operate simulated businesses over extended
periods~\citep{cite_vending}. Such workflows place the model inside a process whose state evolves as it
acts~\citep{cite_memoryarena}. Information gathered through one tool call can determine a later decision,
while an earlier edit, purchase, or commitment can change the options that remain
~\citep{cite_why_reasoning_fails_to_plan}. These workflows therefore extend beyond fully specified question
answering to agentic, interactive tasks. Fig.~\ref{fig:teaser}(a) shows a marked performance gap between the written-out and agentic
forms, while broader evaluations find that reliability declines as the expert time and
interaction horizon required by these workflows grow~\citep{cite_metr_horizon,cite_agent_reliability}.

Meeting this expectation creates a joint environment--signal construction problem. Every training or
evaluation instance needs an executable environment and a dependable signal tied to the outcome produced
within it. Written-out mathematics and reasoning tasks
~\citep{cite_deepseek_r1,cite_open_reasoner_zero,cite_reasoning_gym}, as well as search tasks with answer-based
rewards~\citep{cite_search_r1}, provide cheap, exact signals but present the relevant information in the prompt
rather than require the model to recover it while acting in a changing state. Hand-engineered simulators
~\citep{hubbs2020orgym} restore that interaction and can expose exact outcomes, but specialists must implement
the dynamics and evaluator for each new family~\citep{cite_rlve}. Human-feedback methods
~\citep{cite_christiano_rlhf,cite_instructgpt} reach less structured workflows at the cost of repeated expert
annotation. Learned judges and rubrics~\citep{cite_rubrics_rewards,cite_mockworlds} reduce that burden but
introduce an evaluator whose validity must itself be established~\citep{cite_judge_reliability}. Systems now
generate tools and interactive environments~\citep{cite_autoforge,cite_envscaler,cite_scaleenv,cite_awm},
sometimes together with executable checks~\citep{cite_eigendata}, broadening task coverage and reducing
authoring work. Together, these approaches address only different parts of producing diverse agentic environments
with dependable outcome signals at low extension cost. Yet generative pipelines commonly construct the task or
environment before fixing its reward assignment or evaluation rule, leaving dynamics and evaluation to be aligned afterward.

\begin{figure}[t]
  \centering
  \vspace{-1.0em}
  \includegraphics[width=\linewidth]{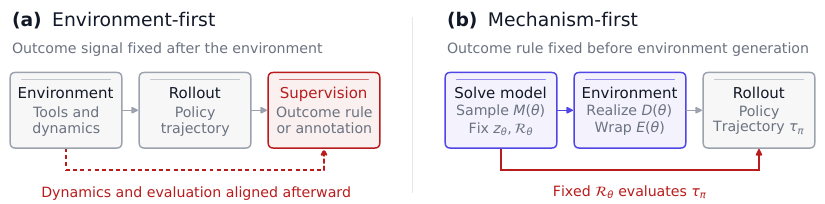}
  \vspace{-2em}
  \caption{Construction order. Environment-first pipelines construct the agent-facing environment $E$ before
  defining its outcome rule or annotating trajectories. \vhd\ first solves a sampled mathematical model $M(\theta)$, whose
  reference outcomes $z_\theta$ fix $\mathcal R_\theta$ before the executable dynamics $D(\theta)$ are realized
  and wrapped as environment $E(\theta)$.}
  \label{fig:inversion}
  \vspace{-1.3em}
\end{figure}

\vhd\ reverses this order. We first sample and solve a mathematical problem before generating the
environment. A frozen setter then uses a sample from a diverse corpus of real-world documents to seed the scenario and implements the
problem's decision process through stateful tools. The sampled problem governs how the environment evolves, while its
solution supplies the outcome signal. The player sees neither the parameters nor the solution and must recover
the relevant information through interaction. Established mathematical model families, such as optimization,
offer mature solvers and a ready source of tasks, while new parameter draws produce additional environments
within each family at low marginal cost. Fig.~\ref{fig:inversion} illustrates the construction order.

The paper makes three contributions. (1) We introduce mechanism-first construction, deriving an agentic
environment and its outcome signal from the same pre-solved problem. (2) We realize it as a corpus-grounded
generation and replay pipeline with information-asymmetric interfaces and automatic admission, producing 3{,}300 admitted environments at
low extension cost. (3) Using these environments to train a Qwen3.6-35B-A3B model raises its mean agentic score across five optimization families
from $0.204$ to $0.815$. The checkpoint improves on all three held-out training and all eight unseen families.
Performance remains strong as the same construction scales tasks and horizons, pointing toward an evolving
training substrate. Externally, it reaches $3.4\times$ the base ending balance and surpasses Qwen3.7-Max on a
365-day storefront~\citep{cite_ecommerce}, improves ten interaction-focused BFCL V4 cells by $2.84$
points~\citep{cite_bfcl_v4}, and
preserves written-out problem solving.

\section{Related Work}
\label{sec:related}

\textbf{Formal and verifiable reasoning.} Written-out mathematics QA pairs each problem with inexpensive
checks~\citep{cite_deepseek_r1,cite_reasoning_gym}, and search agents can likewise receive answer-based
rewards~\citep{cite_search_r1}. These settings score final responses without requiring agents to alter and operate
within a changing task state. SATLM and LLM+P improve reliability by translating problems into logic or planning
languages and delegating inference to a solver~\citep{cite_satlm,cite_llmp}, but still target an answer or plan
rather than an agentic training environment. Formal or human-designed frameworks instead provide exact dynamics
through text games~\citep{cite_textworld}, PDDL environments~\citep{cite_pddlgym}, or language-rendered planning
domains~\citep{cite_autoplanbench}. Procedural generation varies instances under fixed rules
~\citep{cobbe2020procgen,hubbs2020orgym}. Thus verifiability either stops at an answer or plan or retains a domain
specification. \vhd\ keeps the computable reference while using sampled mechanisms and corpus grounding to
produce diverse agentic instances.

\textbf{Generated agentic environments.} As target workflows grow more varied and longer, this per-domain
specification cost becomes harder to sustain. Recent systems reduce it by synthesizing tools and task
environments~\citep{cite_autoforge,cite_envscaler,cite_scaleenv}, or by generating world models and simulators
~\citep{cite_awm,cite_mockworlds,cite_llm_sim}. Across agent training, feedback comes from process or turn-level
rewards~\citep{cite_online_prm,cite_trace,cite_webshepherd}, learned reward models
~\citep{cite_christiano_rlhf,cite_instructgpt}, generated rubrics~\citep{cite_rubrics_rewards}, and executable
verifiers~\citep{cite_eigendata,cite_rlve}. Generation can therefore broaden agency and semantic diversity at
lower authoring cost. Yet learned or generated graders require validation~\citep{cite_judge_reliability}, and
agreement between generated dynamics and evaluation is commonly established afterward~\citep{cite_autoenv}.
Work on partial observability and information gathering~\citep{kaelbling1998pomdp,cite_active_reasoning,cite_proactive_info}
and long-horizon execution~\citep{cite_long_horizon_execution,cite_horizon_mirage} sharpens the behavioral target
but often assumes fixed environments. The two lines therefore cover complementary parts: formal methods secure
outcomes but retain domain authoring, whereas generation reduces authoring but reopens outcome grounding. \vhd\
bridges them by pre-solving each sampled mechanism and retaining its reference for hidden dynamics and graded
scoring.

\section{Formulation}
\label{sec:formulation}

Our goal is to construct diverse agentic environments together with dependable outcome signals, without
repeating the full authoring effort for every instance. The two artifacts form a joint problem because an
environment determines which trajectories can occur, while its evaluator determines what those trajectories are worth. Producing
them as separate artifacts leaves their agreement to be established after generation.

An agentic environment is a process rather than a prompt. Observations depend on state, actions change that
state, and their consequences unfold along a trajectory. We use \emph{dynamics} broadly for the relationships
among states, actions, observations, and outcomes. In a real environment, these dynamics may be unknown and
need not admit an explicit mathematical form. An effective policy nevertheless needs some useful internal proxy
for them, even if it never recovers a set of equations. Operations research often abstracts a concrete decision
problem and its operating conditions into a mathematical model for analysis and policy design
~\citep{hubbs2020orgym}. We take a similar view of agentic environments, treating their behavior as governed
by an underlying model. To construct an environment together with a verifiable outcome signal, \vhd\ reverses
the usual direction. We begin with a mathematical model that usually comes with a verifiable reference solution,
then render its state transitions, information structure, constraints, and objective as a new
environment. The model thereby becomes the common source of the interactive process and
its evaluation.

\subsection{Construction order}

Once an environment $E$ and its outcome rule $\mathcal{R}_{E}$ are fixed, policy learning is conceptually
straightforward. A policy $\pi$ acts in $E$ to produce a trajectory $\tau_\pi$, and $\mathcal{R}_{E}$ maps the
realized outcome to a signal for improving $\pi$. This describes how an environment is used for learning.
Policy optimization itself takes its construction as given. For an existing environment, supervision can be
added either by defining $\mathcal{R}_{E}$ or by assigning an annotation $y_\pi$ to a sampled trajectory
$\tau_\pi$. When the environment itself is generated, it is still commonly constructed before either form of
supervision. We call this order \emph{environment-first}. The interactive process $E$, including whatever transition
structure governs it, is fixed before an outcome rule is defined or trajectories from it are annotated. This
construction need not isolate the transition structure as a separate object, even when that structure is known.

\vhd\ instead makes the upstream mechanism explicit. Let $\theta$ denote sampled parameters and $M(\theta)$
the corresponding mathematical model. Solving the model yields reference outcomes $z_\theta$ and fixes the
outcome rule $\mathcal R_\theta(\,\cdot\,;z_\theta)$. We write $D(\theta)$ for the model's executable realization,
which governs state transitions and utility, and $E(\theta)$ for the agent-facing environment obtained by
wrapping $D(\theta)$ with an interaction interface. For a policy $\pi$, $\tau_\pi$ is the resulting trajectory
and $y_\pi$ its outcome signal. The arrows below record which artifact is available when the next is constructed.
\begin{equation}
\label{eq:orders}
\begin{aligned}
\text{Environment-first}\qquad
&E\xrightarrow{\ \mathrm{define}\ }\mathcal R_E
\quad\text{or}\quad
(E,\pi)\xrightarrow{\ \mathrm{interact}\ }\tau_\pi
\xrightarrow{\ \mathrm{annotate}\ }y_\pi,\\[2pt]
\text{\vhd}\qquad
&M(\theta)\xrightarrow{\ \mathrm{solve}\ }z_\theta
\xrightarrow{\ \mathrm{fix}\ }\mathcal R_\theta(\,\cdot\,;z_\theta),\\[-1pt]
&M(\theta)\xrightarrow{\ \mathrm{realize}\ }D(\theta)
\xrightarrow{\ \mathrm{wrap}\ }E(\theta),\\[-1pt]
&\tau_\pi\sim E(\theta),
\qquad y_\pi=\mathcal{R}_\theta\!\left(\tau_\pi;z_\theta\right).
\end{aligned}
\end{equation}
Environment-first construction may contain rich or even known dynamics. The distinction is that they are
already packaged in $E$ when supervision is supplied. In \vhd, both the executable process and its evaluation
descend from $M(\theta)$. Solving first fixes $z_\theta$ and the corresponding outcome rule before realization
supplies the stateful dynamics and wrapping exposes them to a policy. During training, $E(\theta)$ generates
$\tau_\pi$, while the already fixed $\mathcal R_\theta$ evaluates it against $z_\theta$.
Fig.~\ref{fig:inversion} illustrates this reversal, and App.~\ref{app:factorization-provenance} places common
methods within the same construction view. We next unpack how the construction inherits the useful properties
of the solved mechanism.

\subsection{Model, dynamics, and interface}
\label{ssec:layers}

\begin{figure}[t]
  \centering
  \vspace{-1.2em}
  \includegraphics[width=0.98\linewidth]{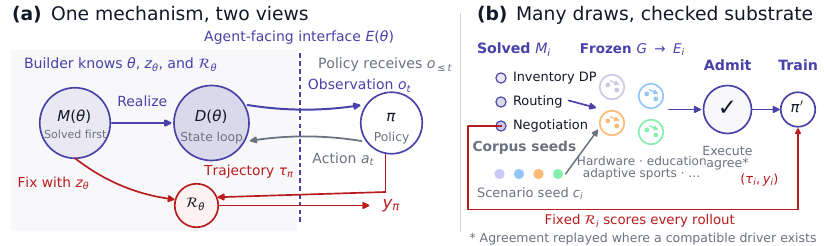}
  \vspace{-0.8em}
  \caption{Mechanism-first construction from one solved instance to a training substrate. (a) The model fixes
  dynamics and evaluation while the interface controls the policy view. (b) Corpus-grounded draws undergo
  execution and supported reference checks before training.}
  \label{fig:layers}
  \vspace{-1.4em}
\end{figure}

The mechanism-first construction separates the mathematical model, its executable dynamics, and the interface
exposed to the policy. A mechanism family $f$ defines a distribution $P_f$ over multi-period decision problems, and a draw
$\theta\sim P_f$ fixes one instance. The mathematical model $M(\theta)$ specifies its decisions, constraints,
objective, and feasible policies. Its executable dynamics $D(\theta)$ contain a state space $\mathcal S$, a
transition $\Lambda_\theta$, and a utility function $U_\theta$. The environment $E(\theta)$ wraps these
dynamics with an observation map $\Omega_\theta$, an action set $\mathcal A$, and a budget of $T$ turns.
Together these objects define the feedback process below, while Fig.~\ref{fig:layers}(a) places the
$M(\theta)\!\to\!D(\theta)\!\to\!E(\theta)$ realization beside the interaction and evaluation paths from the
same model.
\begin{equation}
\label{eq:interface}
\begin{gathered}
D(\theta)=(\mathcal S,\Lambda_\theta,U_\theta),
\qquad
E(\theta)=\operatorname{Wrap}\!\big(D(\theta);\Omega_\theta,\mathcal A,T\big),\\[-1pt]
a_t=\pi(o_{\le t}),
\qquad s_{t+1}=\Lambda_\theta(s_t,a_t),
\qquad o_{t+1}=\Omega_\theta(s_{t+1},a_t),\\[-1pt]
u(\pi;\theta)=U_\theta(\tau_\pi),
\qquad 0\le t<T .
\end{gathered}
\end{equation}

A turn is one answered interface call. A notable feature of agentic environments is that a policy may need to
explore the environment before it can act effectively. Under partial observability and over long horizons, an
agent may spend turns acquiring task-relevant information before making commitments whose effects persist. Let
$\mathcal A_{\mathrm{probe}}\subseteq\mathcal A$ denote these information-acquisition actions. A probe returns
a designated view of hidden state or
parameters, whereas a decision action changes state through $\Lambda_\theta$. Both consume the shared budget
$T$. Probes are not required in every agentic environment, but instantiate the recurring coupling between
gathering information and acting on it. The interface may also provide state-independent computation, which
neither reads nor changes task state.
Withholding parameters yields a partially observed process whose latent state includes the
draw~\citep{kaelbling1998pomdp}. The policy need not reconstruct $M(\theta)$ symbolically, but effective action
requires a useful proxy for the exposed dynamics.

The builder $G$ receives the complete draw and the corpus seed, whereas the policy receives only observations
published through $E(\theta)$:
\begin{equation}
\label{eq:asym}
G\colon (\theta,c)\ \mapsto\ \big(D(\theta),E(\theta)\big)
\qquad\text{whereas}\qquad
\pi\colon o_{\le t}\ \mapsto\ a_t .
\end{equation}
Since $M(\theta)$ and $z_\theta$ already fix the governing structure and evaluation, $G$ is asked to realize
and wrap them rather than invent them jointly. This narrower role reduces the capacity required of the setter.
Sec.~\ref{ssec:synthesis} gives the realization, and App.~\ref{app:setter-capacity} shows that the starting
Qwen3.6-35B-A3B checkpoint, which is later optimized as the player, can already act as the setter and produce
valid agentic environments. A stronger setter primarily improves yield and interface refinement.
The asymmetry lies in what the interface publishes, not in whether the environment depends on $\theta$. Because
$\Omega_\theta$ and $\mathcal A$ belong to the wrapper, they provide an explicit control point over which
components of the draw appear initially, which require interaction to reveal, and which remain latent. In the
agentic form, the default interface offers no direct read of the complete draw or either reference value, so the
policy can obtain only the views returned by its calls. Fig.~\ref{fig:teaser}(a) shows the resulting agentic
difficulty. Models perform strongly on the underlying problems in written-out form but substantially worse in
their stateful agentic form. Because the model and interface are separate,
the same draw can be presented in written-out form without changing the underlying problem or its evaluation.
Resampling $\theta$ and varying its range or horizon produces further instances, while changing the corpus seed
varies their setting and language. These choices provide a controllable information boundary and low-cost
extension once the family-level components exist.

Once $E(\theta)$ is obtained, policy learning shifts from solving a fully specified written-out problem to acting
online through sequential, state-dependent decisions from interface observations, under the same objective and
fixed reference $z_\theta$.
Inventory control gives a concrete example. Per-product demand, one shared capacity, and a joint ordering fee
define $M(\theta)$. The state $s_t$ records stock on hand, prices and a forecast appear through $\Omega_\theta$, and
the demand table remains hidden. Stock bought early occupies capacity later, and no action recovers a lost sale. The
resulting interface therefore retains both information gathering and binding commitments. App.
~\ref{app:families} gives the corresponding objects for all eleven families.

\subsection{Outcome evaluation}
\label{ssec:descent}

The same pre-generation solution fixes how a completed trajectory is evaluated. A family-specific solver
$S_f$ computes $z_\theta=(u^*(\theta),u_0(\theta))$, where $u^*(\theta)$ is the optimal value of $M(\theta)$ and
$u_0(\theta)$ is the value of a fixed default policy. Our realization instantiates the general signal $y_\pi$ as
a scalar episode-level
reward:
\begin{equation}
\label{eq:metric}
S_{f}\colon \theta\ \mapsto\ z_\theta=\big(u^{*}(\theta),\,u_{0}(\theta)\big),
\qquad\qquad
r(\pi;\theta)\;=\;\operatorname{clip}_{[0,1]}\!\left(
\frac{u(\pi;\theta)-u_{0}(\theta)}{u^{*}(\theta)-u_{0}(\theta)}\right).
\end{equation}
The references use closed-form arithmetic, enumeration, or a numerical solver. No language model estimates
or judges either endpoint, and both are fixed before the opening observation. The verified optimum gives the
scale an upper anchor, while the default policy gives it a lower anchor. Eq.~\ref{eq:metric} therefore
maps the default to $r=0$, the full-information optimum to $r=1$, and clips outcomes to this meaningful range
rather than to an arbitrary numerical interval. The score is arithmetic over realized utility and fixed
references, so no learned evaluator enters the scoring path. Under partial information, $u^*(\theta)$ is an upper
reference and need not be attainable by an online policy. Sec.~\ref{ssec:sampling} describes the implementation's
equivalent shift of origin.

\label{ssec:free}
\label{ssec:levels}

Taken together, this construction yields stateful agentic interaction, a verifiable optimum and
reference-grounded score, an explicit information boundary, and diverse instances obtained by resampling
rather than reauthoring.
Sec.~\ref{sec:pipeline} realizes these principles through sampling, solving, corpus-grounded synthesis, and
behavioral admission. Sec.~\ref{sec:exp} then tests whether the resulting
policies improve on new draws, mechanism families, and larger horizons, and whether written-out and agentic
forms expose a distinct operating shortfall.

\section{Producing environments at scale}
\label{sec:pipeline}

Sec.~\ref{sec:formulation} formalizes how a solved mechanism jointly determines the environment dynamics and
its outcome signal. We now turn this principle into a scalable generation pipeline. Fig.~\ref{fig:layers}(b) summarizes the construction, while Tab.~\ref{tab:corpus-overview} reports the resulting substrate.

\noindent\textbf{Sampling and solving.}\label{ssec:sampling} Each instance begins with $\theta\sim P_f$,
followed by the family solver $S_f$. Closed-form procedures,
enumeration, dynamic programming, or numerical optimization compute $z_\theta$ without a language model and
thereby fix the normalized outcome rule in Eq.~\ref{eq:metric}. To make $M(\theta)$ executable, the pipeline maps
its decision variables to state updates, enforces its constraints in the transition function, and accumulates its
objective as terminal utility. The resulting $D(\theta)$ instantiates the same realization logic across parameter
draws, while corpus grounding varies its setting and interface. The inventory example in
App.~\ref{app:example-inventory} shows this mapping concretely, with orders updating stock, capacity limiting
transitions, and revenue and costs determining utility.
Established mathematical and operations-research models make candidate mechanisms straightforward to source
and instantiate. App.~\ref{app:families} catalogues the formulations, decision structures, and reference
procedures of all eleven reported families.

\begin{table}[H]
\vspace{-0.6em}
\centering\small
\setlength{\tabcolsep}{5pt}
\renewcommand{\arraystretch}{1.08}
\begin{tabular}{@{}>{\raggedright\arraybackslash}p{0.19\linewidth}
                    >{\raggedright\arraybackslash}p{0.30\linewidth}
                    >{\raggedright\arraybackslash}p{0.42\linewidth}@{}}
\toprule
\textbf{Dimension} & \textbf{Measured scale} & \textbf{Structure} \\
\midrule
Scenario seeds & \textbf{28} topical domains; no admitted domain above \textbf{4.7\%}
& Real-world documents vary settings, entities, relations, and language \\
Mechanisms & \textbf{3} training families; \textbf{11} evaluated families
& Repeated parameter draws from inventory DP, routing, knapsack, LP/QP, scheduling, search, and related models \\
Environments & \textbf{3,300} admitted; $\geq\!10$ tools each; $\approx$\$0.01--\$0.03 per admitted record
& Stateful dynamics, domain-specific interfaces, and player-facing instructions \\
\bottomrule
\end{tabular}%
\caption{Scale and structure of the generated substrate. Scenario seeds diversify the setting and language,
while mechanism families determine the dynamics and outcome semantics. Apps.~\ref{app:families},
\ref{app:validity}, \ref{app:setter-capacity}, and~\ref{app:corpus} report family definitions, verification
coverage, the cost reconstruction, and full distributions.}
\label{tab:corpus-overview}
\vspace{-1.1em}
\end{table}

\begin{figure}[!t]
  \centering
  \vspace{-1.1em}
  \includegraphics[width=\linewidth]{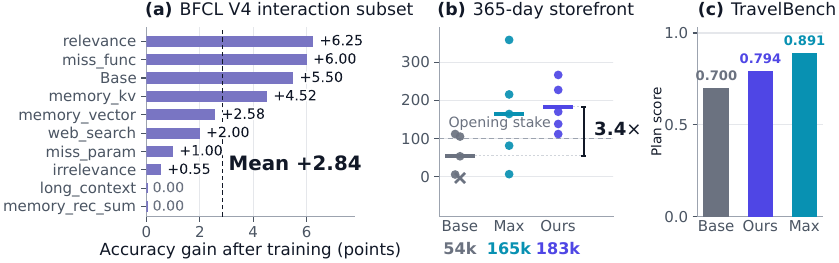}
  \vspace{-1em}
  \caption{External transfer. (a) Ten interaction-focused BFCL V4 cells gain $2.84$ points. (b) Dots show five storefront runs per arm, horizontal segments mark means, and $\times$ marks bankruptcy. The trained checkpoint completes all five and reaches $3.4\times$ the Base mean, above Qwen3.7-Max. (c) TravelBench
  plan quality rises from $0.700$ to $0.794$, below the Qwen3.7-Max reference of $0.891$.}
  \label{fig:external}
  \vspace{-1em}
\end{figure}

\noindent\textbf{Corpus-grounded realization.}\label{ssec:synthesis} For each environment, a frozen language-model setter receives the complete
draw and one independently sampled passage from a diverse corpus of real-world documents, then generates the scenario, relational
database, domain-specific tool schemas and bodies, and player instruction. The passage supplies the entities,
relations, and domain language, while $M(\theta)$ determines the decision process and $z_\theta$ fixes its
evaluation. The Qwen3.6-35B-A3B starting checkpoint already produces valid agent-facing environments in this
role, while a stronger setter mainly improves yield and interface refinement. App.~\ref{app:impl-setter} details
the realization and record structure, and App.~\ref{app:setter-capacity} gives the comparison. Each record retains the complete draw inside $D(\theta)$ and publishes through $E(\theta)$ only the observations and actions defined by its interface. The player therefore acts on state revealed through catalogue, probe, decision, or clock operations rather than reading the latent parameters or references directly. App.~\ref{app:impl-isolation} details this isolation and information boundary.

\noindent\textbf{Admission and scale.}\label{ssec:admission} Admission tests whether a generated candidate
executes and whether its realized behavior agrees with the precomputed construction. Failed candidates are
regenerated or rejected before training. Across the held-out audit, every sampled environment executed
successfully, repeated outcomes agreed, and realized outcomes remained within their stored references.
App.~\ref{app:validity} reports the admission criteria, coverage, and audit statistics. \label{ssec:yield}
The admitted environments draw scenario seeds from 28 topical domains with normalized entropy 0.997. Across them, realizations
range from hardware replenishment and education logistics to adaptive-sports planning. App.~\ref{app:examples}
traces these cases from mechanism to interface and admission, while App.~\ref{app:corpus} reports the complete
environment metadata, family composition, and admission yield. Episodes in these environments average 59.7
assistant turns.

\noindent\textbf{Policy optimization.}\label{ssec:training} Each admitted environment supplies trajectories and
the fixed episode-level signal from Eq.~\ref{eq:metric}.
Our realization optimizes groups of rollouts from the same instance with GRPO~\citep{cite_grpo}.
The mechanism-first construction can be paired with other policy optimizers, as formalized in
App.~\ref{app:factorization-feedback}. App.~\ref{app:impl-training} gives the objective and run configuration, while
App.~\ref{app:protocol} reports the episode, token, and context budgets.

\section{Experiments}
\label{sec:exp}

The experiments address three research questions. \textbf{RQ1.} Do policy gains extend to held-out and unseen mechanisms, external
interfaces, and substantially longer horizons? \textbf{RQ2.} If a model can solve the written-out problem, why
does stateful interaction still leave substantial room for policy improvement? \textbf{RQ3.} Can environment
generation and policy learning co-scale as mechanisms grow?

\vspace{-1mm}
\noindent\textbf{Experimental setup.}\label{ssec:exp-setup} The starting policy is Qwen3.6-35B-A3B (Base). GRPO produces the trained policy. The pipeline produces 3{,}300 admitted environments. Generated-family evaluation covers held-out environments from all three training families and eight
unseen families. Generated-family scores use Eq.~\ref{eq:metric}, while
external benchmarks retain their native metrics. The primary comparison is between the starting and trained checkpoints under the same evaluation protocol.
Qwen3.7-Max provides an additional within-model reference point under its API protocol.
App.~\ref{app:protocol} specifies the sampling, budget, and execution settings.
App.~\ref{app:results} provides the
full result grids and training diagnostics.

\subsection{Training gains and transfer}
\label{ssec:transfer}

\noindent\textbf{Across generated mechanism families.}
At the agentic level, training improves every evaluated family. Fig.~\ref{fig:teaser}(b) organizes eleven
families by what changes from training. Held-out training families gain $+0.56$ on average. Near-OOD families
are unseen optimization families and gain $+0.63$. Far-OOD families change the decision structure or restrict
probing and gain $+0.24$ and $+0.16$ across the two groups. App.~\ref{app:results} reports the complete family grid.

\noindent\textbf{Beyond the generated substrate.}\label{ssec:external}
We evaluate the trained checkpoint on three external agentic or tool-use benchmarks that retain their native
protocols and share neither instances nor interfaces with the generated training environments.
For general function calling, we report BFCL V4's ten Multi-Turn, Agentic, and Hallucination cells, matching our
focus on stateful tool use~\citep{cite_bfcl_v4}. TravelBench asks agents to
use travel tools to assemble multi-day itineraries satisfying coupled timing, budget, commonsense, and
personalized constraints~\citep{cite_travelbench}. E-Commerce Bench runs a 365-day autonomous business in which
an agent manages multiple storefronts, customer orders, returns, market events, and delayed cash
flow~\citep{cite_ecommerce}.

Fig.~\ref{fig:external} summarizes the external results, with the clearest change on the longest-horizon
benchmark. Across five reported storefront runs per arm, the base
checkpoint completes four and goes bankrupt once. The trained 35B-A3B checkpoint completes all five without
bankruptcy over $1{,}147$--$1{,}825$ assistant turns, far beyond the 59.7-turn generated-family average in
Sec.~\ref{ssec:admission}. Its mean ending balance rises from $54{,}294$ to $182{,}844$, or $3.4\times$, and
exceeds the Qwen3.7-Max reference of $165{,}224$. The unweighted mean over this BFCL V4 subset rises from
$61.25$ to $64.08$, and
TravelBench plan quality increases from $0.700$ to $0.794$, below the Qwen3.7-Max reference of $0.891$.
These results show transfer to general tool use, constrained planning, and substantially longer interaction.
App.~\ref{app:results} reports the remaining native metrics and cell-level breakdowns.

\vspace{-2.5mm}
\subsection{A learnable gap under stateful interaction}
\label{ssec:shortfall}

\vspace{-0.6em}
\begin{table}[H]
\vspace{-0.3em}
\centering
\small
\begingroup
\renewcommand{\arraystretch}{1.00}
\setlength{\tabcolsep}{8pt}
\newcommand{\scoremove}[2]{{\color{gray}#1}\,$\rightarrow$\,{\color{vhdpurple}\bfseries #2}}
\begin{tabular}{@{}lccc@{}}
\toprule
& \multicolumn{1}{c}{\scriptsize Complete Problem}
& \multicolumn{1}{c}{\scriptsize Parameters Revealed}
& \multicolumn{1}{c}{\scriptsize Parameters Acquired} \\
\textbf{Family}
& \textbf{Written-Out} $F$
& \textbf{Informed-Agentic} $I$
& \textbf{Agentic} $A$ \\
\midrule
Inventory DP           & \scoremove{0.976}{0.999} & \scoremove{0.469}{0.968} & \scoremove{0.449}{0.912} \\
Knapsack               & \scoremove{0.984}{0.996} & \scoremove{0.132}{0.948} & \scoremove{0.117}{0.843} \\
Linear Programming     & \scoremove{0.978}{1.000} & \scoremove{0.161}{0.732} & \scoremove{0.143}{0.661} \\
Quadratic Programming  & \scoremove{0.975}{1.000} & \scoremove{0.168}{0.836} & \scoremove{0.162}{0.795} \\
Routing                 & \scoremove{0.897}{0.965} & \scoremove{0.226}{0.889} & \scoremove{0.148}{0.866} \\
\midrule
\textbf{Family Macro}  & \scoremove{0.962}{0.992} & \scoremove{0.231}{0.875} & \scoremove{0.204}{0.815} \\
\bottomrule
\end{tabular}
\endgroup
\caption{Five optimization families under written-out ($F$), informed-agentic ($I$), and agentic ($A$)
presentations. Scores show Base $\rightarrow$ trained under the same solver-derived outcome scale.}
\label{tab:forms-main}
\vspace{-1em}
\end{table}

Across five optimization families, we compare each problem in written-out ($F$), informed-agentic ($I$), and
agentic ($A$) forms under the same solver-derived outcome scale. In $F$, the model reasons over the complete
instance and returns an answer, with the Python tool available. In $I$ and $A$, the problem and evaluation stay
fixed, but the model must instead act as a policy through the same multi-step environment. $I$ reveals all
sampled parameters upfront, while $A$ exposes them only through interaction. In inventory control, for example,
$I$ shows the demand for each
product before the episode, whereas $A$ requires the model to collect and analyze demand information before
deciding how much to order.

\noindent\textbf{The construction preserves problem-solving competence while exposing a large, learnable policy gap.}
The base checkpoint scores $0.962$ in $F$ but falls to $0.231$ in $I$, although the sampled parameters remain
available. It must now carry decisions through state changes, shared constraints, and commitments whose
consequences persist across the horizon. Requiring those parameters to be acquired through the interface
further lowers $A$ to $0.204$. Training changes $(F,I,A)$ by $(+0.030,+0.644,+0.611)$, closing $84\%$ of the
parameter-revealed stateful gap and $77\%$ of the full written-out-to-agentic gap.

Representative trajectories in App.~\ref{app:trajectory-evidence} make the gap concrete. In a seven-period allocation task, both checkpoints reveal
all eight items and invoke the code tool. The base policy nevertheless writes ``For each period, find best
combination,'' spends 27 of 29 shared-budget units in the first three periods, and ends with score 0. The trained
policy instead writes ``I need to plan across all 7 periods,'' constructs one horizon-wide
allocation, and reaches score 1. The paired excerpts in App.~\ref{app:traj-horizon} show that the failure is not recognition of the local knapsack but construction of the
coupled horizon-wide model. In the observation case reported in App.~\ref{app:traj-observation}, the base policy states that it needs to probe the items but
acts after revealing only four of eight. The trained policy reveals all eight before planning and again reaches
score 1. These traces expose myopic optimization under persistent cross-period
state and premature commitment under incomplete observation. App.~\ref{app:traj-costly-information} gives a
third paired case in which both policies identify the same leading option, but only the trained policy stops
acquiring information before its cost overwhelms the outcome.

The gain is not explained by only learning to call the code tool. The fixed-stratum analysis in
App.~\ref{ssec:adoption} attributes $+0.030$ of a $+0.443$ agentic gain to increased adoption across the eight
families with both tool-use strata, leaving $+0.413$ within fixed strata. Consistent with a capability-elicitation
interpretation, the gain coincides with a modest policy shift of
$D_{\mathrm{KL}}(\pi_{\mathrm{trained}}\Vert\pi_{\mathrm{Base}})=0.089$ nats per assistant token on held-out
trajectory states, with the estimation protocol and breakdown reported in App.~\ref{app:elicitation}. The construction therefore preserves the mathematical problem while
creating substantial room to learn information acquisition, state-dependent decisions, and sustained execution.
The unseen-family and external gains in Sec.~\ref{ssec:transfer} show how far that policy improvement extends
beyond the original mechanisms and interfaces.

\FloatBarrier
\vspace{-1.5mm}
\subsection{Co-scaling environment and policy frontiers}
\label{ssec:env-frontier}

\begin{wrapfigure}[13]{r}{0.43\linewidth}
  \vspace{-1.7em}
  \centering
  \includegraphics[width=\linewidth]{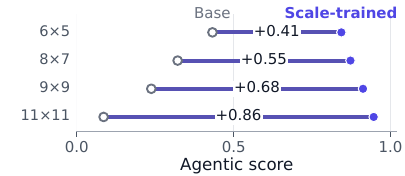}
  \vspace{-2.2em}
  \caption{Environment--policy co-scaling over four knapsack configurations. Each filled point is a separate
  checkpoint trained and evaluated on fresh environments at that scale. Labels give its gain over the common Base.}
  \label{fig:env-frontier}
  \vspace{-0.8em}
\end{wrapfigure}

\noindent\textbf{The construction opens learnable space beyond the current policy.}
RQ3 asks whether environment generation and policy learning can continue to advance as mechanism size and
horizon grow. Fig.~\ref{fig:env-frontier} increases task complexity across four configurations by scaling both
the number of items and the decision horizon, from 6 items over 5 periods to 11 items over 11 periods. We use
two copies of Qwen3.6-35B-A3B. One remains frozen and realizes admitted environments from the sampled
mathematical models and their reference outcomes under a fixed interface, solver, and evaluator. At each scale,
the other starts as Base, trains under a matched budget, and is evaluated on fresh held-out environments at the
corresponding scale. Each filled point represents one such scale-matched checkpoint.

The shared starting checkpoint makes this asymmetry concrete. Across all four scales, its frozen construction
copy continues to realize admitted environments that leave substantial room for its Base policy to improve. At
$6\times5$, the Base and trained policies score $0.432$ and $0.843$. At $11\times11$, they score $0.085$ and
$0.946$. The gains across the sweep are $+0.412$, $+0.551$, $+0.675$, and $+0.861$, with the two larger
configurations averaging $+0.768$ compared with $+0.482$ for the two smaller configurations. The mathematical
model makes the complete structure and reference outcomes available during construction, while the policy
receives only interface observations and must recover the relevant information through interaction. The same
checkpoint can consequently construct an environment that is harder than it can initially navigate as a policy,
and the growing gains show that this opened space remains learnable as mechanism scale increases. This pattern
shows the potential for environment construction and policy learning to co-scale under a shared evaluator.
App.~\ref{app:setter-capacity} reports complementary construction results, and
App.~\ref{app:frontier} gives the exact level-wise values.

\FloatBarrier

\vspace{-3mm}
\section{Conclusion}
\label{sec:conclusion}

Mechanism-first construction turns solved models into diverse stateful environments with solver-derived outcome signals. Corpus grounding supplies the setting, while asymmetric interfaces require information acquisition and consequential action. Across the mechanisms and external benchmarks tested here, the resulting training improves policy performance under changing state while preserving written-out problem solving. The setter and scale experiments further show that a fixed setter can realize new draws and that increasing mechanism size and horizon continues to provide useful training headroom under the same evaluator. By deriving environment dynamics and outcome rules from a shared solved mechanism, \vhd\ extends verifiable reinforcement learning from answers toward trajectories and offers a practical framework for expanding agentic training environments as policy capabilities develop across broader task settings.

\clearpage
\bibliography{biblio}
\bibliographystyle{colm2024_conference}

\clearpage
\appendix
\raggedbottom
{\huge\bfseries Appendix\par}
\vspace{0.45em}
\etocdepthtag.toc{vhdappendix}
\etocsettagdepth{vhdmain}{none}
\etocsettagdepth{vhdappendix}{subsection}
\begingroup
\small
\setlength{\parskip}{4pt}
\renewcommand{\baselinestretch}{1}\selectfont
\makeatletter
\renewcommand*\l@section[2]{%
  \ifnum \c@tocdepth >\z@
    \addpenalty\@secpenalty
    \addvspace{1em \@plus\p@}%
    \setlength\@tempdima{1.5em}%
    \begingroup
      \parindent \z@ \rightskip \@pnumwidth
      \parfillskip -\@pnumwidth
      \leavevmode \bfseries
      \advance\leftskip\@tempdima
      \hskip -\leftskip
      #1\nobreak\hfil
      \nobreak\hb@xt@\@pnumwidth{\hss #2\kern-\p@\kern\p@}\par
    \endgroup
  \fi}
\makeatother
\tableofcontents
\endgroup
\clearpage

\section{Discussion}
\label{sec:discussion}

Scaling agentic training requires diverse settings, realistic stateful interaction, and dependable outcome
signals to grow together. \vhd\ assigns these roles within one construction. Corpus seeds supply settings and
language, the underlying mechanism specifies state transitions and outcome semantics, and agent-facing
interfaces determine what the policy can observe before consequential action. The precomputed reference anchors
the resulting outcome signal.

The comparison across the three presentations exposes a gap between mathematical competence and agentic policy.
The starting checkpoint remains near ceiling when the complete problem is given, yet falls sharply when the same
mechanism unfolds through changing state, shared constraints, and persistent commitments. Most of this gap
remains when the parameters are revealed. Training raises both stateful forms while preserving written-out
problem solving, a pattern consistent with capability elicitation. Representative trajectories connect this
change to information acquisition, cross-period decisions, and the cost of continued exploration.
App.~\ref{app:elicitation} measures the policy shift, and App.~\ref{ssec:adoption} separates it from increased
code-tool use. Improvement also appears on unseen mechanisms and external stateful tasks
(Sec.~\ref{ssec:transfer}).

More broadly, we view the learnable space in agentic training as arising from an asymmetry between the structure
available when an environment is constructed and what a policy can recover through interaction. This asymmetry
may come from human design or the capabilities of a stronger setter. In \vhd, much of the structure behind it
is already supplied by the mathematical model through latent quantities, coupled constraints, and delayed
consequences. The interface determines how this structure is revealed to the policy. The starting 35B
checkpoint can already realize valid new draws, while mechanism parameters extend scale, horizon, and
information cost under the same evaluator. The setter and frontier results suggest that this substrate can
expand with policy competence. App.~\ref{sec:limitations} summarizes the scope of the evidence.

This construction suggests a broader role for formal decision models in agentic learning. Written-out
mathematics has provided language models with a scalable substrate for reasoning because structured problems
admit inexpensive verification~\citep{cite_deepseek_r1,cite_open_reasoner_zero,cite_reasoning_gym}. Agentic
learning requires analogous structure not only in problems and answers, but also in observations, actions, and
consequences across trajectories. Operations research offers a direct route because established models encode
objectives, constraints, resource coupling, and sequential decisions, while mature solution procedures provide
fixed reference outcomes. Their realization and interface turn this structure into an interactive process. We
do not claim that operations research is the only substrate for this transition. More generally, any formal
mechanism that supports parameterized variation, executable realization, controlled information asymmetry, and
an outcome standard fixed before interaction could play a similar role. Operations research is useful here
because established families provide these properties together. The present experiments establish the
feasibility and learnability of this construction over the evaluated families, not that the policy explicitly
reconstructs the underlying model or that operations research is sufficient for agentic competence more
generally.

\FloatBarrier
\section{Admission and verification}
\label{app:validity}

Admission evaluates three criteria during generation and replay. Each can reject a record, subject to the
replay coverage specified below. A failed generation check triggers another attempt. Write $u^{*}(\theta)$ for
the optimal value and $u_{0}(\theta)$ for the default one, both computed by the sampler before the
dynamics $D(\theta)$ existed. Replaying a policy $\pi$ through the generated code accumulates a terminal
cumulative utility $u(\pi;\theta)$. Where the verifier has compatible drivers, it uses up to three:
$\pi^{*}$ solving the family's program, $\pi_{0}$ greedy on the visible values, and $\pi_{\varnothing}$
committing nothing at all. Criteria below appear in the order they apply.

\textbf{C1, executability.} The generated dynamics and declared interface tools must initialize and execute
without exception. Where compatible replay exists, each available reference policy must reach a numeric
terminal outcome.

\textbf{C2, a valid default region.} The default outcome must lie in a family-defined admissible region
appropriate to the objective's scale and direction. A candidate outside this region is regenerated.

\textbf{C3, reward agreement at the optimum.}
For replay-supported families, admission requires the reference driver to reproduce the precomputed optimum
within a family-specific numerical tolerance. Routing uses a separate driver that acts through the environment
interface and applies the same agreement check to the optimum and default outcomes.

A stratified post-generation audit drew ten held-out environments from each of the eight replay-supported
families and ran oracle, default, and idle policies twice from fresh state. Every family passed all ten
environments. Across 480 executions, none raised an exception, all 240 repeated terminal outcomes agreed, and
no replayed raw utility exceeded its precomputed $u^*(\theta)$. App.~\ref{app:examples} shows the corresponding
mechanism, interface, and outcome structure for inventory, routing, and knapsack.

Post-generation replay selects a driver from the interface roles declared by each record rather than from its
family label. The shared driver covers seven reported optimization families, namely inventory DP, facility
location, budget-coupled knapsack, linear and quadratic programming, scheduling, and scheduling with tardiness.
It executes a family solution obtained in closed form, by enumeration, by dynamic programming, or from a
numerical solver, then compares the realized outcome with the precomputed reference.

Routing uses the separate direct-interface replay described under C3. The three other reported families also
have computable references. Negotiation uses backward induction to obtain a session oracle. Sequential stopping
and sequential search use hindsight ceilings together with their default references. These quantities enter the
outcome rule directly. What varies across families is how the reference is obtained and whether the shared
driver can execute a corresponding trajectory, not whether a reference exists.

\section{Implementation and record structure}
\label{app:implementation}

\subsection{Reference computation}
\label{app:impl-reference}

The eleven reported families compute their references by a closed-form procedure, exact enumeration, dynamic
programming, or numerical optimization. Inventory
replenishment uses backward Bellman recursion over joint inventory states. Routing uses Held-Karp recursion
over served-customer subsets, facility location enumerates open-facility subsets, and tardiness scheduling
applies exponential recursion over job subsets. Linear programming uses HiGHS and the separable quadratic
programme uses SLSQP through \texttt{scipy.optimize}. No language model is called during reference computation.

All later components receive the same parameter dictionary and reference pair.

\paragraph{Family registration.} A mechanism family is registered through a parameter sampler, reference
procedure, and reusable realization or replay adapter. The reported families adapt established mathematical and
operations-research formulations and solver routines for these components. Once registered, the same components
serve repeated parameter and corpus draws, while environment generation and admission remain automated.

\subsection{Setter passes}
\label{app:impl-setter}

The setter receives an independently sampled real-world source passage, its topical label, and the
complete parameter vector. It emits an entity and relationship model, a backing database, at least ten
executable tools, the player instruction, an executable default policy, and the required output format.
Candidates that fail execution or interface alignment are refined or regenerated before admission. The source
corpus spans 28 topical domains, and each environment is grounded in a separately sampled passage.

Reusable family dynamics are supplied by the sampler where available, while the setter instantiates dynamics
for families without that reusable implementation. The reference pair remains sampler-computed in every
environment, regardless of which component wrote the dynamics.

\begin{table}[t]
\centering\small\setlength{\tabcolsep}{4.5pt}
\begin{tabular}{llrrrr}
\toprule
\textbf{setter} & \textbf{family} & \textbf{emitted} & \textbf{replay} & \textbf{attempts} & \textbf{refined} \\
\midrule
\multirow{3}{*}{Qwen3.6-35B-A3B}
 & inventory DP & 10/10 & 9/10 & 2.70 & 40\% \\
 & routing & 8/10 & 8/8 & 2.00 & 50\% \\
 & tardiness & 4/10 & 4/4 & 3.00 & 50\% \\
\midrule
\multirow{3}{*}{Qwen3.7-Max}
 & inventory DP & 10/10 & 10/10 & 1.10 & 0\% \\
 & routing & 10/10 & 10/10 & 1.00 & 0\% \\
 & tardiness & 10/10 & 10/10 & 1.00 & 0\% \\
\bottomrule
\end{tabular}
\caption{Setter-capacity feasibility check over three mechanism families. Attempts are averaged over emitted
records, and refined is the share whose tool interface was revised during synthesis. The comparison uses the
same families but not identical parameter draws.}
\label{tab:setter-ablation}
\end{table}

\subsection{Setter capacity}
\label{app:setter-capacity}
In Tab~\ref{tab:setter-ablation}, we test whether mechanism-first realization itself requires a frontier setter by using a frozen copy of the
Qwen3.6-35B-A3B starting checkpoint, without environment-specific training, on ten requested records from each
of inventory DP, routing, and tardiness scheduling. We compare its aggregate generation and verification
outcomes with existing Qwen3.7-Max records from the same families. The parameter draws are not paired, so the
comparison measures feasibility and realization burden rather than a causal model effect.

The starting checkpoint emits 22 of 30 requested records, and 21 of its 22 emitted environments preserve the
reference under compatible replay, giving an effective admission rate of 70\%. Qwen3.7-Max emits and passes all
30 records, giving 100\%, while reducing the mean number of attempts from 2.50 to 1.03 and the
interface-refinement rate from 45.5\% to zero. The smaller setter therefore establishes feasibility, while the
stronger setter improves generation efficiency and interface alignment.

These additional attempts erase the smaller setter's per-token price advantage in the available reconstruction.
At the lowest public rates we found, both setters cost roughly $\$0.01$--$\$0.03$ per admitted environment, and
Qwen3.6-35B-A3B is not cheaper per admission despite its lower token rate.\footnote{The lowest rates found were
$\$0.05/\$0.70$ per million input/output tokens for Qwen3.6-35B-A3B on
\url{https://openrouter.ai/qwen/qwen3.6-35b-a3b}, and \textyen6/\textyen18 for the Qwen3.7-Max real-time alias on
\url{https://help.aliyun.com/zh/model-studio/model-pricing}. Prices were accessed August 28, 2026. The Max rate
is a limited 50\%-off promotion, and the dollar conversion uses \textyen7 per US dollar.}

\subsection{Environment and interpreter isolation}
\label{app:impl-isolation}

Each environment retains the complete sampled draw and reference pair inside its executable dynamics. The
player-facing interface exposes only the declared catalogue, probe, commitment, and clock operations, without
access to latent values or references. Every presentation provides the same state-independent computation
utility, which cannot access environment internals. The written-out and agentic forms therefore differ in
parameter access rather than in access to computation.

\subsection{Training implementation}
\label{app:impl-training}

Training uses GRPO with 64 prompts per step and 16 retained rollouts per prompt, oversampled to 18 attempts.
The advantage is the return minus its group mean, without standard-deviation normalization, a KL penalty, or a
value model. One inner epoch consumes each batch at learning rate $2\times10^{-6}$ and clip ratio
$4\times10^{-3}$. The reported checkpoint follows 34 optimizer steps, consumes 2{,}176 environments from the
2{,}200-environment training partition, and produces 34{,}816 scored rollouts.

Tab.~\ref{tab:settings} reports the resource and optimization settings.

\FloatBarrier
\section{Representative generated environments}
\label{app:examples}

\begin{figure}[H]
  \centering
  \includegraphics[width=\linewidth]{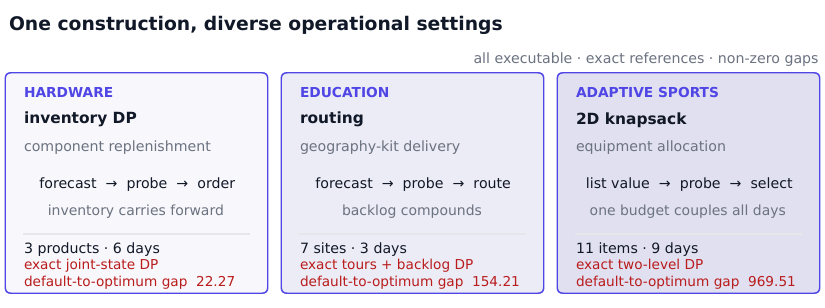}
  \caption{Representative generated environments across three corpus domains and mechanism families. Each panel
  shows the information-acquisition and commitment pattern exposed to the policy, the state carried across the
  horizon, and the exact solver-backed advantage over the fixed default.}
  \label{fig:examples}
\end{figure}

This section traces three admitted records from their corpus setting and sampled mechanism through the generated
interface and fixed evaluation. Two belong to the training split and one to an unseen evaluation family.
Fig.~\ref{fig:examples} shows how inventory control, routing, and knapsack become operational tasks in hardware,
education, and adaptive sports. Aggregate statistics in Tab.~\ref{tab:corpus-overview} characterize diversity
across the admitted set, while these cases make the mechanism-to-interface mapping concrete.

Tab.~\ref{tab:example-validation} records the reference procedure, raw default-to-optimum gap, and admission
evidence for each example. All three have a positive gap, and compatible replay reproduces the reference through
the generated interface.

\begin{table}[H]
\centering\fontsize{7}{8}\selectfont\setlength{\tabcolsep}{3.5pt}
\begin{tabular}{@{}p{2.05cm}p{3.55cm}p{3.10cm}p{1.45cm}p{2.75cm}@{}}
\toprule
\textbf{mechanism} & \textbf{reference computation} & \textbf{raw default $\rightarrow$ optimum} & \textbf{gap} & \textbf{admission evidence} \\
\midrule
inventory DP & backward DP over joint inventory states & $22.74\rightarrow45.01$ & $22.27$ & executable; reference replay passed \\
routing & Held--Karp tours with DP over backlog & $-117.81\rightarrow36.40$ & $154.21$ & executable; reference replay passed \\
two-dimensional knapsack & per-period knapsack DP with DP over the shared budget & $371.45\rightarrow1{,}340.96$ & $969.51$ & executable; reference replay passed \\
\bottomrule
\end{tabular}
\caption{Solver and admission evidence for the representative records. The positive raw gaps establish that
the instances are not solved by reproducing the fixed default policy.}
\label{tab:example-validation}
\end{table}

\subsection{Inventory control as hardware replenishment}
\label{app:example-inventory}

The inventory example is grounded in the hardware domain and rendered as six days of replenishment at the
fictional Shenzhen Bay Hardware Distribution Center. Its generated instruction asks the agent to manage an
Industrial Servo Motor, a Precision Optical Encoder, and a Thermal Management Unit. The visible description
states that one truck carries at most six units per day, an order of any size incurs a joint setup cost of
$37.69$, each product can hold and receive at most five units, unsold inventory carries forward, and unmet
demand incurs a lost-sales penalty. This is operational language for one sampled joint-replenishment model,
rather than an independently authored reward.

\begin{vhdprompt}{Generated instruction excerpt---hardware inventory}
\small
You are the operations director at the Shenzhen Bay Hardware Distribution Center, managing the
replenishment of three critical automation components: an Industrial Servo Motor, a Precision Optical
Encoder, and a Thermal Management Unit. The planning horizon spans six days.

Each day, you decide how many units of each component to order. A single shared freight truck carries at
most six units across all three products, and dispatching it incurs a joint setup cost of $37.69$ whenever
any quantity is ordered. Each product can receive at most five units per day and has storage capacity five.
Unsold inventory carries forward with a holding cost, while unmet demand incurs a lost-sales penalty.

True demand is deterministic but hidden. Forecast stickers are visible but may be misleading. Probing one
SKU reveals its complete demand trajectory across all six days. Maximize cumulative profit by balancing
information gathering, shared truck capacity, joint setup costs, inventory carryover, and lost sales.
\end{vhdprompt}

Tab.~\ref{tab:example-inventory} gives the model's forecast and hidden demand trajectories. Each tuple in the
economics column gives unit price, order cost, holding cost, and lost-sales penalty.

\begin{table}[H]
\centering\footnotesize\setlength{\tabcolsep}{4pt}
\begin{tabular}{@{}lccc@{}}
\toprule
\textbf{product} & \textbf{economics} & \textbf{visible forecast} & \textbf{hidden demand} \\
\midrule
Servo Motor X12 & $(9.37,2.61,0.89,0.46)$ & $(5,5,5,2,4,1)$ & $(6,5,4,2,4,2)$ \\
Optical Encoder Z8 & $(12.05,3.13,1.07,0.30)$ & $(1,5,2,5,1,2)$ & $(2,3,3,4,2,2)$ \\
Thermal Unit T5 & $(11.21,3.72,1.19,1.62)$ & $(3,5,5,5,2,5)$ & $(3,6,3,5,4,5)$ \\
\bottomrule
\end{tabular}
\caption{The sampled quantities underlying the hardware inventory example. The forecast is available through
the interface; probing a product reveals its full hidden demand trajectory.}
\label{tab:example-inventory}
\end{table}

Concretely, the sampler renders order quantities as action inputs, inventory balance as persistent state,
capacity and setup costs as transition constraints, and the objective terms as accumulated utility. The generated
interface exposes catalogue and inventory views, a per-product demand probe, a replenishment action, a day-advance
action, and helpers for capacity, holding cost, setup cost, and horizon planning. The dynamics update inventory
only after an order and day advance, charge the joint setup cost once on any ordering day, carry unsold stock
forward, and make lost sales irreversible. The sampler computes a raw default value of
$22.74$ and a raw optimum of $45.01$; after the implementation's shift of origin, the stored reference pair is
$u_0=0$ and $u^*=22.27$. The inventory reference driver reproduces the solver-derived outcome through the
generated interface.

\subsection{Routing as education logistics}
\begin{vhdprompt}{Generated instruction excerpt---education routing}
\small
You coordinate the Cascade Mountains Geography Kit Delivery Program. A single vehicle leaves a central
depot to serve seven remote learning sites over three days, carrying at most 20 educational kits each day.
Every dispatch is an ordered route whose cost is determined by the known road network.

True demand at each learning site is hidden behind a static forecast sticker. Probing a site reveals its
complete demand trajectory across the horizon. Demand not served on its arrival day becomes backlog and
incurs a penalty of $1.60$ per unit for every day it remains outstanding, while each delivered kit earns
$4.23$. Maximize cumulative realized value by deciding which sites to probe, which ordered route to dispatch,
and how to manage limited capacity against compounding backlog.
\end{vhdprompt}

The generated interface supports probing, route dispatch, capacity and distance checks, backlog accounting,
site profiles, and horizon planning. The exact solver combines Held--Karp routing with dynamic programming over
backlog. Its raw default and optimum are $-117.81$ and $36.40$, yielding a non-trivial gap of $154.21$. The
candidate executes successfully, and its reference policy reproduces the optimum through the generated tools.

\subsection{Knapsack allocation as adaptive-sports planning}
\begin{vhdprompt}{Generated instruction excerpt---adaptive-sports allocation}
\small
You are the equipment manager at a wheelchair-basketball training center preparing for a regional qualifier.
Across nine consecutive days, you allocate eleven equipment items to a display shelf with capacity 17. A
single procurement budget of 76 must cover the full horizon, so spending on an early day reduces the budget
available later.

Each item's size and stocking cost are known, but its true tournament revenue varies deterministically by day
and is hidden behind a static list value. Probing one item reveals its full revenue trajectory. Each day you
must choose a feasible subset, load the shelf, and advance the environment. Maximize cumulative savings by
balancing daily capacity, information acquisition, and the shared multi-day budget.
\end{vhdprompt}

The generated interface supports item inspection, revenue probing, shelf-feasibility checks, subset selection,
budget tracking, and day advancement. An exact two-level dynamic program computes the raw default and optimum,
$371.45$ and $1{,}340.96$, giving a non-trivial gap of $969.51$. The candidate executes successfully, and its
reference policy reproduces the optimum through the generated tools.

\FloatBarrier
\section{Construction, feedback, and optimization}
\label{app:factorizations}

Sec.~\ref{sec:formulation} separates the provenance of an interactive task from the feedback used to train
on it. The separation yields four axes, summarized in Tab.~\ref{tab:construction_axes}. Environment
construction determines where the dynamics and interface originate. Outcome provenance determines which
object assigns meaning to successful behavior. Feedback localization determines where that assessment is
attached to a trajectory. Policy optimization determines how the resulting quantities update a model. The
four choices need not move together. Together with Eq.~\ref{eq:orders}, they form a common map. A method is
located by the source of its environment, the provenance and placement of its feedback, and the optimizer that
consumes that feedback.

\begin{table}[H]
\centering\footnotesize\setlength{\tabcolsep}{3pt}
\begin{tabular}{@{}p{2.35cm} p{2.85cm} p{3.40cm} p{2.00cm} p{1.90cm}@{}}
\toprule
\raggedright\textbf{Construction} & \raggedright\textbf{Environment source} &
\raggedright\textbf{Outcome provenance} & \raggedright\textbf{Localization} &
\raggedright\textbf{Optimizer} \tabularnewline
\midrule
\raggedright Environment-first & \raggedright existing or generated dynamics and interface &
\raggedright fitted reward model, rubric model, or judge & \raggedright prefix, turn, or terminal &
\raggedright independent choice \tabularnewline
\raggedright Environment-first & \raggedright existing or generated task or program &
\raggedright authored or generated checker attached to the task & \raggedright usually terminal &
\raggedright independent choice \tabularnewline
\raggedright Mechanism-first & \raggedright realization generated from sampled $\theta$ &
\raggedright references computed from the same $\theta$ before realization &
\raggedright independent choice, terminal here & \raggedright independent choice, GRPO here \tabularnewline
\bottomrule
\end{tabular}
\caption{Four axes that are often conflated in descriptions of agent training. The representative rows show
that feedback localization and policy optimization can change without changing the provenance from which an
environment and its outcome were constructed.}
\label{tab:construction_axes}
\end{table}

\subsection{Construction order and method coverage}
\label{app:factorization-provenance}

Within this map, the principal distinction introduced by \vhd\ is the construction order in
Eq.~\ref{eq:orders}. Its arrows record which artifact
is available when the next is constructed. Along the environment-first branch, $E$ exists before either
$\mathcal R_E$ is defined or sampled trajectories receive annotations $y_\pi$. Along the \vhd\ branch, solving
$M(\theta)$ first yields $z_\theta$ and fixes $\mathcal R_\theta$. Realization then produces $D(\theta)$ and
$E(\theta)$. The mechanism guides realization, while the precomputed standard provides a reference for checking
it. The provenance, verification, and extension differences in Tab.~\ref{tab:mechanism-envfirst} follow from
this reversal.

For an environment $E$ that is already available, a learned evaluator uses
$\hat r_{\psi}(\tau)=J_{\psi}(E,\tau)$, with validity resting on the fitted model or judge
~\citep{cite_christiano_rlhf,cite_instructgpt,cite_judge_reliability}. An executable verifier uses
$r(\tau)=V_E(\tau)$, where $V_E$ may be authored or generated after the task, while a sampled trajectory may
instead receive an annotation $y_\pi$. These choices change the provenance and placement of feedback while
retaining $E$ as the upstream object.

Rubric-based rewards refine the learned-evaluator case. A rubric supplies criteria $q_1,\ldots,q_m$, a grader
assigns criterion scores $J_{\psi,j}(E,\tau,q_j)$, and an aggregation rule forms
$\hat r_{\mathrm{rubric}}(\tau)=\sum_{j=1}^{m}w_jJ_{\psi,j}(E,\tau,q_j)$
~\citep{cite_rubrics_rewards}. This changes the structure and coverage of feedback while its semantics still
come from the rubric, grader, and aggregation rule introduced for $E$. Thus outcome type and feedback
localization can vary without changing the construction dependency in Eq.~\ref{eq:orders}. Learned reward or
judge pipelines, rubric-based supervision, trajectory annotation, and executable verifiers
are therefore represented by different settings of the same four axes. Mechanism-first construction adds the
alternate upstream branch in Eq.~\ref{eq:orders}. The change concerns construction provenance rather than
feedback localization or policy optimization.

\subsection{A released environment-first comparison}
\label{app:mechanism-envfirst}

EnvScaler provides a strong public instance of the environment-first branch in Eq.~\ref{eq:orders}
~\citep{cite_envscaler}. It first generates a reusable program skeleton containing state, rules, and tools,
then instantiates its database and task. This yields the executable environment $E$, whose released training
scenarios contain many API-based interactions. EnvScaler subsequently generates the checklist and final-state
validators that define the scenario-specific outcome rule $\mathcal R_E$. Its construction therefore follows
$E\rightarrow\mathcal R_E$ in our definition. Both substrates provide stateful interaction, diverse APIs, and
computed rewards. \vhd\ instead fixes its objective and references before realizing the environment.
Tab.~\ref{tab:mechanism-envfirst} compares the two orders directly.

\begin{table}[H]
\centering\footnotesize
\setlength{\tabcolsep}{3.5pt}
\renewcommand{\arraystretch}{1.10}
\begin{tabular}{@{}>{\raggedright\arraybackslash}p{0.15\linewidth}
                    >{\raggedright\arraybackslash}p{0.37\linewidth}
                    >{\raggedright\arraybackslash}p{0.37\linewidth}@{}}
\toprule
\textbf{Dimension} & \textbf{Mechanism-first (\vhd)} & \textbf{Environment-first (EnvScaler)} \\
\midrule
\textbf{Construction order}
& Solve $M(\theta)$ to fix $z_\theta$ and $\mathcal R_\theta$ before realizing $D(\theta)$ and wrapping $E(\theta)$.
& Generate a stateful program skeleton, database, and task, then derive the checklist and final-state validators. \\

Dynamics and APIs
& The mechanism fixes transition semantics. The setter renders corpus-grounded relational data and executable APIs.
& An LLM generates the state schema, business rules, and Python tools. \\

Outcome signal
& Realized terminal utility is normalized between a default and a precomputed reference from the same mechanism.
& Reward is the fraction of LLM-generated Boolean checklist functions satisfied by the final state. \\

Verification
& Reference replay compares generated dynamics with outcomes computed before realization.
& AST checks and a 100-round testing--checking agent loop filter skeletons by execution behavior. \\

Information boundary
& The builder sees the draw and references. The player sees interface views, and three forms vary this boundary.
& State is hidden behind tools. A conversation mode reveals task information through a simulated user. \\

Extension unit
& Register a sampler, reference procedure, and reusable realization once per family. Then resample parameters and corpus seeds or increase size and horizon.
& Generate databases, tasks, and validators within a skeleton. A new skeleton adds discovery, synthesis, and assessment. \\

Reusable setup
& Standard mathematical formulations and solution procedures are already available. Integrating a family requires one reusable sampler and realization/replay adapter.
& Each program skeleton is generated and assessed. EnvScaler reports approximately \$1.024 per skeleton~\citep{cite_envscaler}. \\

Human input
& Once a family is registered, draws, realization, and admission require no per-environment labels.
& The main generation loop uses no per-scenario human labels. Manual judgments are reported for comparison with a model-based quality audit; their scale and labor are unspecified. \\

Marginal cost
& Approximately \$0.01--\$0.03 per admitted environment under the reconstruction in App.~\ref{app:corpus}.
& \$0.0635 per scenario, or approximately \$0.08--\$0.09 after amortizing reported skeleton generation~\citep{cite_envscaler}. \\

\bottomrule
\end{tabular}
\caption{Construction-order comparison. Both substrates provide executable state, diverse APIs, and computed
terminal feedback. EnvScaler constructs the executable substrate and task before generating its validators.
\vhd\ fixes the outcome standard before realization, yielding different provenance, verification, and extension
paths.}
\label{tab:mechanism-envfirst}
\end{table}

Among the systems reviewed in Sec.~\ref{sec:related}, EnvScaler's combination of executable skeletons, task
scenarios, validator code, an RL corpus, training infrastructure, and trained checkpoints makes it one of the
most complete public environment-first baselines available~\citep{cite_autoforge,cite_envscaler,cite_scaleenv}.

Tab.~\ref{tab:mechanism-envfirst} compares the closest marginal units, an EnvScaler scenario and a \vhd\
admitted environment. Both pipelines aim to replace recurring human
trajectory labels with executable signals. EnvScaler further calibrates a model-based quality audit against
manual judgments, while \vhd\ checks generated behavior against precomputed references.

To compare the resulting training substrates under a common learner, we adapt 2{,}200 released EnvScaler RL
scenarios drawn across 51 program skeletons while preserving their native environments and checklist rewards.
The control starts from the same Qwen3.6-35B-A3B checkpoint and uses the same GRPO optimizer, batch size,
rollout group size, and 34-step budget as \vhd. The released RL scenarios form its training substrate, and step
34 is fixed before external evaluation. Benchmark instances and inference settings are then shared across the
two trained checkpoints. BFCL, TravelBench, and E-Commerce Bench are external to both training corpora.

\begin{figure*}[t]
  \centering
  \includegraphics[width=0.9\textwidth]{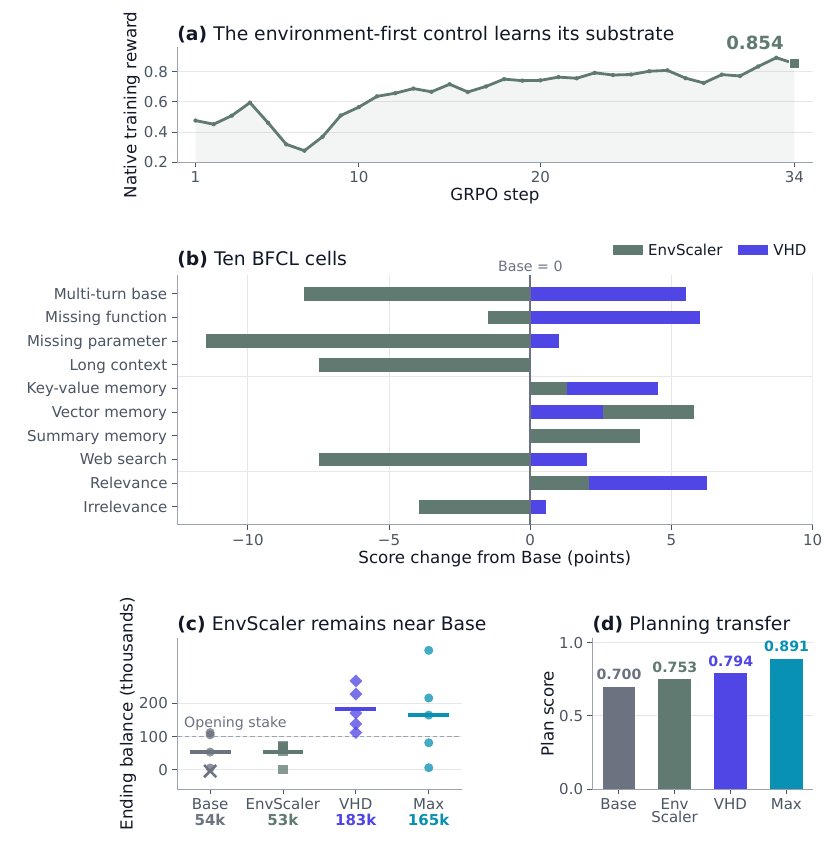}
  \caption{Matched environment-first control. Panel (a) reports EnvScaler's native training reward over 34
  GRPO steps. Panel (b) reports change from Base on ten BFCL cells. Panel (c) shows five 365-day storefront
  runs per model, with horizontal segments marking means and crosses marking bankruptcy. Panel (d) reports
  TravelBench plan quality.}
  \label{fig:envscaler-control}
\end{figure*}

EnvScaler's original experiments show substantial gains for smaller policies from lower starting scores. Its
Qwen3-8B SFT-and-RL checkpoint raises BFCL-v3 Multi-Turn from 28.88 to 41.88~\citep{cite_envscaler}. Our matched
control evaluates a different operating point with a stronger 35B starting policy. Training raises its native
reward over 34 steps, improves TravelBench plan quality from 0.700 to 0.753, and brings all five storefront runs
to day 365. Aggregate transfer is less uniform. Its BFCL ten-cell mean is 58.56 against 61.25 for Base, while
mean storefront balance remains close to Base at 52{,}661 versus 54{,}294. Under the same 35B starting checkpoint
and update budget, \vhd\ improves all three external aggregates. Fig.~\ref{fig:envscaler-control} and
Tab.~\ref{tab:envscaler-control} summarize the comparison. The two regimes suggest that the large gains from a
lower 8B baseline do not automatically translate into uniform gains when continuing from a stronger 35B policy.

\begin{table}[H]
\centering\scriptsize
\setlength{\tabcolsep}{3.5pt}
\renewcommand{\arraystretch}{1.04}
\begin{tabular}{@{}llrrr@{}}
\toprule
\textbf{Benchmark} & \textbf{Cell or measure} & \textbf{Base} & \textbf{EnvScaler} & \textbf{\vhd} \\
\midrule
\multirow{10}{*}{BFCL V4}
 & Multi-turn base          & 65.00 & 57.00 & \textbf{70.50} \\
 & Missing function         & 44.50 & 43.00 & \textbf{50.50} \\
 & Missing parameter        & 50.00 & 38.50 & \textbf{51.00} \\
 & Long context             & \textbf{55.50} & 48.00 & \textbf{55.50} \\
 & Key-value memory         & 60.00 & 61.29 & 64.52 \\
 & Vector memory            & 65.16 & \textbf{70.97} & 67.74 \\
 & Summary memory           & 59.35 & \textbf{63.23} & 59.35 \\
 & Web search               & 57.00 & 49.50 & \textbf{59.00} \\
 & Relevance                & 75.00 & 77.08 & \textbf{81.25} \\
 & Irrelevance              & 80.97 & 77.04 & \textbf{81.52} \\
\cmidrule(lr){2-5}
 & \textbf{Ten-cell mean}   & 61.25 & 58.56 & \textbf{64.08} \\
\midrule
TravelBench & Plan score    & 0.700 & 0.753 & \textbf{0.794} \\
\midrule
\multirow{3}{*}{Storefront}
 & Mean ending balance      & 54{,}294 & 52{,}661 & \textbf{182{,}844} \\
 & Day-365 completion       & 4/5 & \textbf{5/5} & \textbf{5/5} \\
 & Bankruptcies             & 1/5 & \textbf{0/5} & \textbf{0/5} \\
\bottomrule
\end{tabular}
\caption{Matched environment-first control. BFCL reports all ten interaction-focused cells. Both trained
checkpoints start from the same base model and use the same GRPO setting, 2{,}200 training records, and
34-step update budget.}
\label{tab:envscaler-control}
\end{table}

\FloatBarrier

\subsection{Feedback and optimization}
\label{app:factorization-feedback}

Mechanism-first construction determines the provenance of an environment and its evaluative standard. It does
not prescribe when feedback is attached to a trajectory or how that feedback updates a policy. Once the
construction from $M(\theta)$ has supplied $E(\theta)$, $z_\theta$, and $\mathcal R_\theta$, the downstream
choices can be written as
\begin{equation}
\begin{gathered}
y_\pi^{\mathrm{term}}=\mathcal R_\theta(\tau_{\pi,\leq T};z_\theta),
\qquad
y_{\pi,t}^{\mathrm{proc}}=\mathcal R_{\theta,t}(\tau_{\pi,\leq t};z_\theta),\\[-1pt]
\pi^{+}=\operatorname{Update}(\pi;\tau_\pi,y_\pi).
\end{gathered}
\end{equation}
Here $y_\pi$ denotes either the terminal scalar or the sequence of process signals.
Outcome models, process reward models, turn-level checks, and generated verifiers instantiate different
choices of $\mathcal R$ and its input~\citep{cite_gsm8k,cite_prm_verify,cite_online_prm,cite_turn_reward}.
The update may likewise use PPO, GRPO, or another optimizer~\citep{cite_ppo,cite_grpo}. The reported realization
sets $y_\pi$ to the normalized scalar terminal reward in Eq.~\ref{eq:metric} and instantiates
$\operatorname{Update}$ with GRPO. App.~\ref{app:impl-training} gives the corresponding objective and run
configuration.

\FloatBarrier
\section{Evaluation settings}
\label{app:protocol}

Tab.~\ref{tab:settings} reports the resource and policy-optimization settings. We set the maximum episode length
to 1{,}000 turns, the per-turn generation cap to 16{,}384 tokens, the cumulative generation cap to 120{,}000
tokens, the prompt cap to 8{,}192 tokens, and the managed context window to 65{,}536 tokens.
We count one sampled mechanism, its realized interface, and its fixed outcome rule as one environment instance.
The reported set contains 3{,}300 such environments. Of these, 2{,}200 form the training partition, 300 provide
held-out evaluation for the three training families, and 800 provide evaluation for the eight unseen families.
Each evaluation family contributes 100 environments.
Across the trained checkpoint's agentic generated-family exports, episodes have median $31$, mean $59.7$, and
95th percentile $211$ assistant turns.

\begin{table}[H]
\centering\small\setlength{\tabcolsep}{6pt}
\begin{tabular}{@{}p{0.55\linewidth}p{0.34\linewidth}@{}}
\toprule
\textbf{Setting} & \textbf{Value} \\
\midrule
\multicolumn{2}{@{}l}{\textit{Resource Settings}} \\
\addlinespace[1pt]
Maximum Turns per Episode & 1{,}000 \\
Generation Cap per Turn & 16{,}384 tokens \\
Maximum Cumulative Generation & 120{,}000 tokens \\
Prompt Cap & 8{,}192 tokens \\
Managed Context Window & 65{,}536 tokens \\
Observed Assistant Turns, Median / Mean / P95 & 31 / 59.7 / 211 \\
\addlinespace[3pt]
\multicolumn{2}{@{}l}{\textit{Policy Optimization}} \\
\addlinespace[1pt]
Algorithm & GRPO \\
Prompts per Step & 64 \\
Rollouts per Prompt & 16 retained, oversampled to 18 \\
Learning Rate & $2\times10^{-6}$ \\
Clip Ratio & $4\times10^{-3}$ \\
Epochs per Batch & 1 \\
KL Penalty & none \\
Advantage Standardization & none \\
Reported Checkpoint & step 34 \\
\bottomrule
\end{tabular}
\caption{Resource and policy-optimization settings used in the reported runs.}
\label{tab:settings}
\end{table}

\FloatBarrier
\section{Mechanism families}
\label{app:families}

Tab.~\ref{tab:families} catalogues the eleven families the paper draws instances from. Each row gives
the readable name, the mathematical problem its sampler draws, the decision structure an episode presents,
the procedure that computes $u^{*}(\theta)$, and the evaluation band. Across the optimization families the default policy behind $u_{0}(\theta)$
follows one recipe. Act on the sticker numbers, split any shared budget evenly over the horizon, never
probe, and let the true parameters settle what that earns. The linear and the quadratic rows call the same
numerical routine as their reference optimum, one period at a time on the visible coefficients, while
facility location enumerates open subsets on the advertised serving costs. The reference low end is
therefore a solved decision taken on misleading data. Outside that group the default is set per family.
Bargaining concedes ten percent per round, sequential stopping
commits to the first sealed option, and sequential search inspects everything before claiming the best of
what it saw.

\begin{table}[t]
\centering\fontsize{6.5}{7.4}\selectfont\setlength{\tabcolsep}{3pt}
\begin{tabular}{@{}p{1.75cm} p{2.95cm} p{3.45cm} p{2.85cm} p{1.65cm}@{}}
\toprule
\raggedright\textbf{family} & \raggedright\textbf{mathematical problem} & \raggedright\textbf{decision structure} & \raggedright\textbf{reference optimum computed by} & \raggedright\textbf{band} \tabularnewline
\midrule
\raggedright inventory DP & \raggedright multi-product capacitated joint replenishment & \raggedright agentic horizon, inventory carries between periods, one shared order capacity, joint setup charged once per ordering period, demand revealed by probe & \raggedright backward dynamic program over the joint integer inventory state & \raggedright in-domain held out \tabularnewline
\raggedright routing & \raggedright multi-period capacitated vehicle routing with backlog & \raggedright agentic; one vehicle leaves the depot each period and tours the chosen subset, unserved demand carries as backlog and is charged, demand revealed by probe & \raggedright exact Held-Karp tour per subset, with a dynamic program over the backlog carry & \raggedright in-domain held out \tabularnewline
\raggedright negotiation & \raggedright sequential bilateral bargaining under a cash ledger & \raggedright agentic; about 25 deals against a counterpart kernel, bankroll carries, per-deal round budget, bankruptcy at zero & \raggedright backward induction per deal, then the best attainable session total & \raggedright in-domain held out \tabularnewline
\raggedright knapsack & \raggedright multi-period two-dimensional 0/1 knapsack with a shared budget & \raggedright agentic; a shelf-space cap and a stocking-cost cap each period, one procurement budget over the whole horizon, probe reveals an item's profit trajectory & \raggedright two-dimensional knapsack dynamic program per period, plus a dynamic program over the budget split & \raggedright near-OOD \tabularnewline
\raggedright linear programming & \raggedright multi-period linear program with a shared budget & \raggedright agentic; constraint matrix disclosed, drifting objective coefficients hidden and revealed by probe & \raggedright the stacked all-periods program solved by HiGHS & \raggedright near-OOD \tabularnewline
\raggedright quadratic programming & \raggedright multi-period separable concave quadratic program with a shared budget & \raggedright agentic; curvature disclosed, drifting linear coefficients hidden and revealed by probe & \raggedright sequential quadratic programming on the stacked coupled program, global under concavity & \raggedright near-OOD \tabularnewline
\raggedright sequential stopping & \raggedright optimal stopping over sealed options & \raggedright strictly sequential; reveal the next option or commit the last revealed one, and a passed option cannot be revisited & \raggedright the largest value in the sequence, a hindsight ceiling no online policy attains & \raggedright far-OOD, distinct decision structure \tabularnewline
\raggedright sequential search & \raggedright Weitzman search with per-option inspection costs & \raggedright inspect any option at its own charge, then claim one, and every inspection is paid for & \raggedright the best single inspection chosen in hindsight, again not attainable online & \raggedright far-OOD, distinct decision structure \tabularnewline
\raggedright facility location & \raggedright uncapacitated facility location, one instance per period & \raggedright agentic with a probe budget of six tenths of the items; opening costs and demands disclosed, serving costs hidden and drifting, periods independent & \raggedright enumeration over open subsets, each customer assigned to its cheapest open facility & \raggedright far-OOD, probe-budgeted \tabularnewline
\raggedright scheduling & \raggedright single machine, weighted completion time & \raggedright agentic with a probe budget; processing times disclosed, weights hidden and drifting, a full permutation committed each period & \raggedright Smith's rule, a sort by weight over processing time, provably optimal & \raggedright far-OOD, probe-budgeted \tabularnewline
\raggedright scheduling (tardiness) & \raggedright single machine, weighted tardiness, NP-hard & \raggedright agentic with a probe budget; processing times and due dates disclosed, weights hidden and drifting & \raggedright exact dynamic program over job subsets & \raggedright far-OOD, probe-budgeted \tabularnewline
\bottomrule
\end{tabular}
\caption{Every mechanism family the paper uses, with the mathematics it samples, the decision
structure an episode presents, the procedure that computes the reference optimum, and the evaluation band.
Sequential stopping and sequential search are scored against a hindsight bound no online policy
attains, so their rows compare the two arms rather than measure distance to an attainable optimum.}
\label{tab:families}
\end{table}

\FloatBarrier
\section{Additional experimental results}
\label{app:results}

\subsection{Training dynamics}
\label{ssec:run}

\begin{figure}[t]
  \centering
  \includegraphics[width=\linewidth]{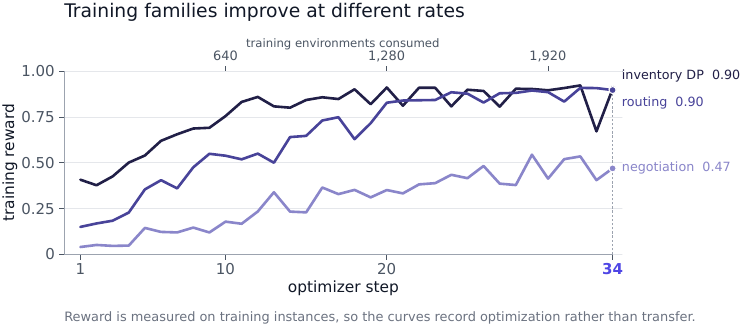}
  \caption{Per-family training reward through the reported checkpoint. These values are measured on training
  instances and document optimization rather than transfer; held-out evaluation provides the transfer evidence.}
  \label{fig:curve}
\end{figure}

Fig.~\ref{fig:curve} plots its per-family rewards for the reported training families against
optimizer step. Inventory DP climbs from $0.408$ to $0.904$ and routing from $0.150$ to
$0.898$, while negotiation reaches $0.470$ from a near-zero start.

The curve establishes only that optimization proceeded. Reward is defined on generated environments, so its
rise cannot distinguish learning a stateful policy from fitting training instances. Step $34$ supplies every
trained-model number in the paper, while held-out evaluation provides the transfer evidence.

\subsection{Generated-family results}

Fig.~\ref{fig:teaser}(a) summarizes the three-arm written-out-to-agentic comparison; the corresponding
family-level cells appear in Tab.~\ref{tab:gap}. Tab.~\ref{tab:family_grid} gives the complete base-to-trained
grid behind the transfer bands in Fig.~\ref{fig:teaser}(b). Reading its agentic columns downward recovers the
per-family results of Sec.~\ref{ssec:transfer}, while reading across a row compares written-out and agentic
forms.

\begin{table}[H]
\centering\scriptsize\setlength{\tabcolsep}{4pt}
\begin{tabular}{lcccccc}
\toprule
& \multicolumn{2}{c}{\textbf{Base}} & \multicolumn{2}{c}{\textbf{\vhd\ (step 34)}}
& \multicolumn{2}{c}{\textbf{Qwen3.7-Max}} \\
\cmidrule(lr){2-3}\cmidrule(lr){4-5}\cmidrule(lr){6-7}
\textbf{family} & written & agentic & written & agentic & written & agentic \\
\midrule
inventory DP           & 0.976 & 0.449 & 0.999 & \textbf{0.912} & 1.000 & 0.489 \\
knapsack               & 0.984 & 0.117 & 0.996 & \textbf{0.843} & 1.000 & 0.525 \\
linear programming     & 0.978 & 0.143 & 1.000 & \textbf{0.661} & 1.000 & 0.399 \\
quadratic programming  & 0.975 & 0.162 & 1.000 & \textbf{0.795} & 1.000 & 0.588 \\
routing                & 0.897 & 0.148 & 0.965 & \textbf{0.866} & 0.996 & 0.142 \\
\midrule
\textbf{mean}          & 0.962 & 0.204 & 0.992 & \textbf{0.815} & 0.999 & 0.429 \\
\textbf{shortfall} & \multicolumn{2}{c}{$0.758$} & \multicolumn{2}{c}{$\mathbf{0.177}$}
& \multicolumn{2}{c}{$0.571$} \\
\bottomrule
\end{tabular}
\caption{Five optimization families in written-out and agentic form, scored by
Eq.~\ref{eq:metric}. Written out minus agentic defines the shortfall. Training reduces the mean shortfall from
$0.758$ to $0.177$ while retaining near-ceiling written-out performance. Bold marks the better of the base and
trained checkpoints.}
\label{tab:gap}
\end{table}

\begin{table}[H]
\centering\small\setlength{\tabcolsep}{3pt}
\begin{tabular}{lccccc}
\toprule
& \multicolumn{2}{c}{written out ($F$)} & \multicolumn{2}{c}{agentic ($A$)} & \\
\cmidrule(lr){2-3}\cmidrule(lr){4-5}
\textbf{family} & Base & \vhd & Base & \vhd & $\Delta A$ \\
\midrule
inventory DP           & 0.976 & 0.999 & 0.449 & 0.912 & $+0.463$ \\
knapsack               & 0.984 & 0.996 & 0.117 & 0.843 & $+0.726$ \\
linear programming     & 0.978 & 1.000 & 0.143 & 0.661 & $+0.518$ \\
quadratic programming  & 0.975 & 1.000 & 0.162 & 0.795 & $+0.633$ \\
routing                & 0.897 & 0.965 & 0.148 & 0.866 & $+0.718$ \\
facility location      & 1.000 & 1.000 & 0.390 & 0.646 & $+0.256$ \\
scheduling             & 1.000 & 1.000 & 0.522 & 0.607 & $+0.085$ \\
scheduling (tardiness) & 1.000 & 1.000 & 0.390 & 0.537 & $+0.147$ \\
negotiation            & n/a   & n/a   & 0.047 & 0.555 & $+0.509$ \\
sequential stopping    & n/a   & n/a   & 0.350 & 0.767 & $+0.417$ \\
sequential search      & n/a   & n/a   & 0.310 & 0.376 & $+0.066$ \\
\midrule
\textbf{mean, five-family subset} & 0.962 & 0.992 & 0.204 & \textbf{0.815} & $\mathbf{+0.611}$ \\
\bottomrule
\end{tabular}
\caption{Every reported generated-family result for the base model and trained checkpoint. Written out exposes
all parameters in the prompt and permits Python-tool use, whereas agentic form exposes them through the environment. Negotiation,
sequential stopping and sequential search were evaluated only in agentic form, so their written-out cells are
absent rather than zero. The mean row covers the five families evaluated in both forms.}
\label{tab:family_grid}
\end{table}

\paragraph{Capability-elicitation diagnostic.}
\label{app:elicitation}
To complement the three-presentation results in Tab.~\ref{tab:forms-main}, we estimate
$D_{\mathrm{KL}}(\pi_{\mathrm{trained}}\Vert\pi_{\mathrm{Base}})$ on 96 held-out early- and
mid-trajectory states from four environment-frontier levels, scheduling, and search. Sampling up to 512
assistant tokens from the trained checkpoint at temperature one and scoring the same tokens under Base gives a
token-weighted forward KL of $0.089$ nats per token, with a task-bootstrap 95\% interval of $[0.082,0.097]$.
The early- and mid-trajectory estimates are $0.150$ and $0.048$, respectively, and the six stratum estimates
range from $0.080$ to $0.095$.

\FloatBarrier
\subsection{Trajectory evidence}
\label{app:trajectory-evidence}
Tab.~\ref{tab:trajectory-evidence} draws three paired cases from held-out generated-family evaluations. They
expose incomplete observation at the interface, myopic planning under cross-period state, and continued
information acquisition after its expected value no longer justifies its cost.

\begin{table}[H]
\centering
\small
\renewcommand{\arraystretch}{1.12}
\setlength{\tabcolsep}{4pt}
\begin{tabular}{@{}>{\raggedright\arraybackslash}p{0.20\linewidth}
                    >{\raggedright\arraybackslash}p{0.36\linewidth}
                    >{\raggedright\arraybackslash}p{0.36\linewidth}@{}}
\toprule
\textbf{Failure pattern} & \textbf{Trace from Base} & \textbf{Trained contrast} \\
\midrule
Myopic cross-period model
& After revealing all eight items, writes ``For each period, find best combination,'' spends 27 of 29
shared-budget units in the first three periods, and ends with score 0.
& Writes ``I need to plan across all 7 periods,'' computes one horizon-wide allocation, and
ends with score 1. \\
\addlinespace
Incomplete observation
& Writes ``Need to probe items to learn true profits,'' reveals four of eight, then begins committing actions and scores 0.
& Reveals all eight trajectories before planning, completes seven periods, and scores 1. \\
\addlinespace
Costly over-exploration
& Identifies the eventual choice after two low-cost inspections but continues through all 14 candidates, pays
134 in total, and scores 0.
& Stops after eight inspections costing 27, commits to the same choice, and scores 0.799. \\
\bottomrule
\end{tabular}
\caption{Illustrative paired trajectories rather than estimates of failure prevalence. Quoted phrases are
verbatim excerpts. Action sequences are compressed from the execution records.}
\label{tab:trajectory-evidence}
\end{table}

The excerpts below retain the policy's own wording while omitting repeated tool payloads.

\subsubsection{Horizon-wide planning}
\label{app:traj-horizon}

The first pair separates local optimization from horizon-wide planning. Both policies reveal all eight items
and invoke the code tool, but Base decomposes the task by period and spends most of its shared budget early.
The trained policy solves and then executes one coupled plan.

\begin{vhdprompt}{Local optimization versus a horizon-wide plan}
\small
\textbf{Base, score 0.} ``For each period, find best combination within space=13 and some budget.''

\textbf{Trained, score 1.} ``The optimal strategy is found. Let me save it and start executing.'' Its stored plan
specifies all seven period-level selections under the shared budget before the first commitment.
\end{vhdprompt}

\subsubsection{Information acquisition before commitment}
\label{app:traj-observation}

In the second case, Base recognizes both the need to inspect and the value of a late-period item, but
commits before completing observation and exhausts the shared budget before that item peaks. The trained policy
instead completes observation and solves one allocation over the full horizon.

\begin{vhdprompt}{Incomplete observation and premature commitment}
\small
\textbf{Base, score 0.} ``Item 7 is very valuable in later periods! Let me update memory and stock the shelf for
period 0.'' Later, ``Budget only 4 left. Can't afford item 7.''

\textbf{Trained, score 1.} ``Now let me probe all 8 items to learn their true profit trajectories.'' After
collecting them, ``Now let me use dynamic programming to find the optimal strategy across all 7 periods.''
\end{vhdprompt}

\subsubsection{Stopping costly information acquisition}
\label{app:traj-costly-information}

The third pair isolates the value of further information. Both policies identify the same leading option after
the first few inspections. Base continues through all 14 candidates, pays 134 in inspection costs, and
finishes with net value $-40.71$. The trained policy stops after eight inspections costing 27 and commits to the
same option for net value $66.29$.

\begin{vhdprompt}{Information value versus inspection cost}
\small
\textbf{Base, score 0.} ``Tsaritsa Couture (index 7) is leading with 93.29 impact. Let me scout a few more to
ensure I'm not missing a better option.'' It eventually scouts every candidate before selecting index 7.

\textbf{Trained, score 0.799.} ``Any additional scouting would only reduce my net value unless I find something
with an impact score dramatically higher than 93.29, which seems unlikely.'' It stops and selects index 7.
\end{vhdprompt}

\subsection{External-benchmark details}

Fig.~\ref{fig:external} summarizes the main external results. Tab.~\ref{tab:external} gives the
three-benchmark summary, and Tab.~\ref{tab:bfcl_full} gives all ten reported function-calling cells and their
changes.

\begin{table}[H]
\centering\footnotesize\setlength{\tabcolsep}{4pt}
\begin{tabular}{llccc}
\toprule
\textbf{benchmark} & \textbf{measure} & \textbf{Base} & \textbf{\vhd} & change \\
\midrule
BFCL V4 interaction subset
 & unweighted mean, ten cells     & 61.25 & \textbf{64.08} & $+2.84$ \\
\midrule
travel planning
 & plan score                     & 0.700 & \textbf{0.794} & $+0.094$ \\
\midrule
365-day storefront
 & mean ending balance            & 54{,}294 & \textbf{182{,}844} & $3.4\times$ \\
\bottomrule
\end{tabular}
\caption{Three external benchmarks under their native metrics. The BFCL row is the unweighted mean over the ten
interaction-focused cells.}
\label{tab:external}
\end{table}

\begin{table}[H]
\centering\small\setlength{\tabcolsep}{6pt}
\begin{tabular}{llccc}
\toprule
\textbf{category} & \textbf{subset} & \textbf{Base} & \textbf{\vhd\ (step 34)} & $\Delta$ \\
\midrule
\multirow{4}{*}{multi-turn}
 & base                    & 65.00 & 70.50 & $+5.50$ \\
 & missing function        & 44.50 & 50.50 & $+6.00$ \\
 & missing parameter       & 50.00 & 51.00 & $+1.00$ \\
 & long context            & 55.50 & 55.50 & $+0.00$ \\
\midrule
\multirow{4}{*}{agentic}
 & key-value memory        & 60.00 & 64.52 & $+4.52$ \\
 & vector memory           & 65.16 & 67.74 & $+2.58$ \\
 & recursive-summary memory& 59.35 & 59.35 & $+0.00$ \\
 & web search              & 57.00 & 59.00 & $+2.00$ \\
\midrule
\multirow{2}{*}{hallucination}
 & relevance               & 75.00 & 81.25 & $+6.25$ \\
 & irrelevance             & 80.97 & 81.52 & $+0.55$ \\
\midrule
\textbf{average} & & 61.25 & \textbf{64.08} & $\mathbf{+2.84}$ \\
\bottomrule
\end{tabular}
\caption{The ten interaction-focused BFCL V4 cells, Base against the reported checkpoint. Eight improve and two
are unchanged. The displayed average is unweighted, and its change is computed before rounding.}
\label{tab:bfcl_full}
\end{table}

Completed checkpoint storefront runs average $1{,}359$ assistant turns, and TravelBench comprises 20 English
tasks.

\subsection{Code-tool adoption control}
\label{ssec:adoption}

\begin{figure}[H]
  \centering
  \includegraphics[width=0.90\linewidth]{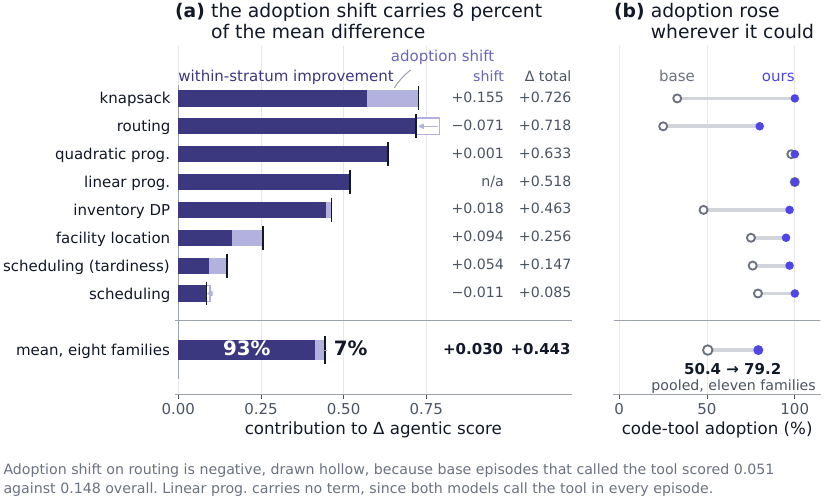}
  \caption{Code-tool adoption control. Panel (a) decomposes score gains, and panel (b) compares adoption rates.}
  \label{fig:adoption}
\end{figure}

Fig.~\ref{fig:adoption} tests whether the agentic gain is explained by increased code-tool use.

Panel (a) uses an Oaxaca--Blinder decomposition to separate each base-to-trained score gain into increased
code-tool adoption and improvement within fixed tool-use strata. Under this descriptive decomposition, most of the observed score difference appears within fixed tool-use strata rather than through the change in tool-call frequency. Routing has a negative adoption term because
its code-using base episodes score below its overall mean. Panel (b) shows that pooled adoption across eleven
evaluation families rises from 50.4 to 79.2 percent. Most of the gain therefore comes from improvement
conditional on tool use rather than from calling the tool more often.

\FloatBarrier
\section{Admitted-environment statistics}
\label{app:corpus}

\begin{figure}[t]
  \centering
  \includegraphics[width=\linewidth]{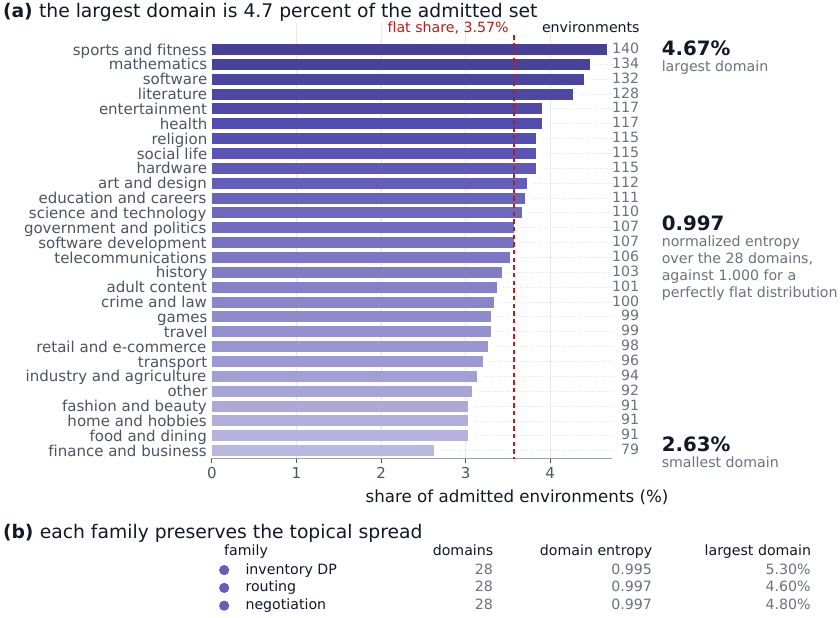}
  \caption{Composition of the three-family generation pool used to form the training partition. Panel (a) shows
  the 28 topical domains associated with the sampled corpus passages, with the dashed line marking a uniform
  split. Panel (b) reports the same composition by mechanism family. Normalized Shannon entropy,
  $H(p)/\log 28$, is 0.997 overall and at least 0.995 within each family.}
  \label{fig:corpus}
\end{figure}

Tab.~\ref{tab:yield} reports the partition sizes, admission yield, reference margins, token use, and reconstructed
per-admission cost. Fig.~\ref{fig:corpus} characterizes the three-family generation pool used to form the
training partition, with cuts by topical domain and mechanism family. Its 28 topical domains have normalized
entropy 0.997 overall and at least 0.995 within each family.

\begin{table}[H]
\centering\small\setlength{\tabcolsep}{4pt}
\begin{tabular}{lc}
\toprule
\textbf{Property of Admitted Environments} & \textbf{Value} \\
\midrule
Reported Environments                                             & 3{,}300 \\
Policy-Optimization Partition                                     & 2{,}200 across three families \\
In-Domain Held-Out Evaluation                                     & 300 (100 per training family) \\
Unseen-Family Evaluation                                          & 800 (100 per unseen family) \\
Setter Attempts per Admission, Mean / Median / Maximum            & 1.41 / 1 / 5 \\
Admitted on the First Setter Attempt                              & 65.6\% \\
Tool Interface Rewritten                                          & 31.6\% \\
Topical Labels Represented                                        & 28 \\
Largest Label Share, Pooled                                       & 4.67\% \\
Normalized Label Entropy                                          & 0.997 \\
Relative Margin, Median / Tenth Percentile                        & 1.00 / 0.34 \\
Input Tokens per Admission                                        & $\approx 900$ \\
Output Tokens per Admission                                       & $\approx 6{,}300$ \\
Public-Rate Cost per Admission                                    & $\approx \$0.01$--$\$0.03$ \\
\bottomrule
\end{tabular}
\caption{Metadata summary of the admitted environments. Token counts and public-rate prices are aggregate
per-admission reconstructions. Setter-capacity details appear in App.~\ref{app:setter-capacity}.}
\label{tab:yield}
\end{table}

\FloatBarrier
\section{Environment-complexity sweep}
\label{app:frontier}
Scale labels denote items $\times$ decision periods in the knapsack family.
Tab.~\ref{tab:env-frontier} gives the exact
level-wise values plotted in Fig.~\ref{fig:env-frontier}.

\begin{table}[H]
\centering\small\setlength{\tabcolsep}{7pt}
\begin{tabular}{lccccc}
\toprule
\multicolumn{3}{c}{\textbf{environment scale}} & \multicolumn{3}{c}{\textbf{agentic score}} \\
\cmidrule(lr){1-3}\cmidrule(lr){4-6}
\textbf{level} & \textbf{items} & \textbf{periods} & \textbf{Base} & \textbf{\vhd} & \textbf{gain} \\
\midrule
L1 & 6  & 5  & 0.432 & 0.843 & $+0.412$ \\
L2 & 8  & 7  & 0.321 & 0.872 & $+0.551$ \\
L3 & 9  & 9  & 0.237 & 0.912 & $+0.675$ \\
L4 & 11 & 11 & 0.085 & 0.946 & $+0.861$ \\
\bottomrule
\end{tabular}
\caption{Scale-matched training as the size and horizon of the knapsack family increase. Each \vhd\ cell is a
separate checkpoint trained at that level and evaluated on fresh held-out environments of the same scale.}
\label{tab:env-frontier}
\end{table}

\FloatBarrier
\section{Limitations and scope}
\label{sec:limitations}

All reported families have computable reference outcomes. Optimization families use exact or numerical solvers,
negotiation uses backward induction, and sequential stopping and sequential search use hindsight ceilings. The
latter provide consistent upper anchors and need not be attainable by an online policy. Post-generation replay
checks generated dynamics against precomputed outcomes over oracle, default, and idle paths on eight families.
App.~\ref{app:validity} reports the held-out audit and its coverage.

The reported implementation trains on one terminal outcome, so the policy optimizer receives no intermediate
supervision. Training covers three families, and generated-family evaluation covers eleven. These are mostly
well-established models from mathematics and operations research whose formulations and solution procedures are
available independently of our pipeline (App.~\ref{app:families}).

Generated-family results use one training run and one evaluation seed. External results measure transfer
separately under each benchmark's native protocol. The reported ablations characterize presentation, tool use,
and scale. The $F/I/A$ comparison isolates how information exposure changes evaluation difficulty, while
training uses the complete construction and does not separately attribute gains to the outcome rule and
interface.

\end{document}